\documentclass[journal,10pt,twoside]{IEEEtran}

\usepackage{verbatim}
\usepackage[pdftex]{graphicx}
\usepackage{epsfig,subfigure}
\usepackage{amsmath,amssymb,amsfonts,bm,amscd} 
\usepackage{algpseudocode}
\usepackage{mathtools} 
\usepackage{mathrsfs} 
\usepackage{amsthm} 
\usepackage{setspace}
\usepackage{fancyhdr}
\usepackage{algorithm} 
\usepackage{psfrag}
\usepackage{cite}
\usepackage{makecell}
\usepackage{url}
\usepackage{float}
\usepackage{soul}
\usepackage{color}
\usepackage{multirow}

\usepackage[table]{xcolor}

\usepackage{color, xcolor}
\usepackage{etoolbox}    

\newtheorem{remark}{Remark}

\newtheorem{assumption}{\textbf{Assumption}}
\newtheorem{definition}{\textbf{Definition}}

\newcommand{\rr}{\mathbb{R}}

\newcommand{\efficiencycruise}[0]{$ {\text{MEAN}}$  
                        & $ {\text{MAX}}$      
                        & $ {\text{MIN}} $ }
\newcommand{\efficiencycomputation}[0]{ $ {\text{MEAN}}$  
                        & $ {\text{MAX}}$      
                        & $ {\text{MIN}} $}
\newcommand{\rom}[1]{(\expandafter{\romannumeral #1\relax})}  
\makeatletter
\pretocmd\@bibitem{\color{black}\csname keycolor#1\endcsname}{}{\fail}
\newcommand\citecolor[1]{\@namedef{keycolor#1}{\color{blue}}}
\makeatother  
 
\begin{document}
\title{ 
{Occlusion-Aware Contingency Safety-Critical Planning for Autonomous Driving }
 
}  
\author{ 
Lei Zheng, Rui Yang, Minzhe Zheng, Zengqi Peng, Michael Yu Wang, \textit{Fellow, IEEE,} and Jun Ma
\thanks{This work was supported in part by the Guangdong Basic and Applied Basic Research Foundation under Grant 2025A1515011812; and in part by the Guangzhou Education Bureau  Project under Grant 2024312095.\textit{(Corresponding author: Jun Ma.)} } 
    \thanks{Lei Zheng, Rui Yang, Minzhe Zheng, and Zengqi Peng are with the Robotics and Autonomous Systems Thrust, The Hong Kong University of Science and Technology (Guangzhou), Guangzhou 511453, China (e-mail: \{lzheng135, ryang253, mzheng615, zpeng940\}@connect.hkust-gz.edu.cn).}%
    \thanks{Michael Yu Wang is with the School of Engineering, Great Bay University, Dongguan 523000, China (e-mail: mywang@gbu.edu.cn).}%
\thanks{Jun Ma is with the Robotics and Autonomous Systems Thrust, The Hong Kong University of Science and Technology (Guangzhou), Guangzhou 511453, China, and also with the Cheng Kar-Shun Robotics Institute, The Hong Kong University of Science and Technology, Hong Kong SAR, China (e-mail: jun.ma@ust.hk). }
} 
\markboth{IEEE TRANSACTIONS ON CYBERNETICS, VOL. 56, NO. 5, May 2026}
{Zheng \MakeLowercase{\textit{et al.}}:  OCCLUSION-AWARE SAFETY-CRITICAL PLANNING}

\maketitle 

\begin{abstract}
Ensuring safe driving while maintaining travel efficiency for autonomous vehicles in dynamic and occluded environments is a critical challenge. This paper proposes an occlusion-aware contingency safety-critical planning approach for real-time autonomous driving. Leveraging {reachability analysis} for risk assessment, forward reachable sets of {phantom vehicles} are used to derive risk-aware dynamic velocity boundaries.
These velocity boundaries are incorporated into a biconvex nonlinear programming (NLP) formulation that formally enforces safety using spatiotemporal barrier constraints, while simultaneously optimizing exploration and fallback trajectories within a receding horizon planning framework. 
 To enable real-time computation and coordination between trajectories, we employ the consensus alternating direction method of multipliers (ADMM) to decompose the biconvex NLP problem into low-dimensional convex subproblems. The effectiveness of the proposed approach is validated through simulations and real-world experiments in occluded intersections. Experimental results demonstrate enhanced safety and improved travel efficiency, enabling real-time safe trajectory generation in dynamic occluded intersections under varying obstacle conditions. The
project page is available at \url{https://zack4417.github.io/oacp-website/}.
\end{abstract}

\begin{IEEEkeywords}
Safety-critical planning, autonomous vehicles, trajectory optimization, alternating direction method of multipliers.
\end{IEEEkeywords}
      
\section{Introduction}
 \IEEEPARstart{A}{utonomous} vehicles (AVs) are increasingly recognized as a transformative solution for urban mobility~\cite{chen2022milestones, guan2022integrated,hou2024risk}. {To realize this vision}, AVs must efficiently plan safe and feasible trajectories in dynamic environments~{\cite{lyu2019multivehicle,chai2021multiobjective,hentout2024shortest}}. Despite significant advancements, addressing potential hazards and enhancing travel efficiency remain critical challenges, especially in environments like urban intersections~{\cite{peng2025,peng2024reward,zhang2023adaptive}}. One of the key factors contributing to this challenge is the presence of occlusions,  where sensors such as cameras and LiDAR may be obstructed by buildings or surrounding vehicles (SVs)~\cite{park2023occlusion,van2024overcoming}. In such scenarios, the ego vehicle (EV) must continuously evaluate occlusion risks and rapidly replan its trajectory in response to unforeseen contingencies, ensuring safety in changing and unpredictable traffic~{\cite{xing2024online,zhang2024automated}.} Contingencies, defined as potential events that cannot be predicted with certainty~\cite{alsterda2021contingency}, require the EV to account for the movements of {phantom vehicles (PVs)} when planning safe trajectories.  
Moreover, real-time trajectory replanning in such environments involves addressing intricate constraints and objectives simultaneously~\cite{Bono2022,yu2023command,chai2021multiobjective, van2024overcoming,he2025model}.  This process inherently introduces non-convexity and nonlinearity into the optimization problem, making real-time replanning computationally intensive~\cite{zheng2025barrier}. While existing approaches, such as game-theoretic methods~\cite{Zhan-RSS-21, qiu2024inferring} and partially observable Markov decision processes~\cite{9564424,zhang2024occlusion}, offer promising solutions for occlusion-aware safe scenarios, they are computationally intractable as the problem size increases. These challenges are further exacerbated in dense and interaction-heavy traffic environments~\cite{zheng2024}.  

In general, receding horizon planning and control are foundational strategies for safe and efficient autonomous driving in dynamic environments~\cite{huang2022learning,shi2021advanced,zheng2024}. These methods incorporate real-time feedback and anticipate future events, enabling AVs to adapt to changing traffic conditions and interact safely with SVs~\cite{lyu2024safety}. To further handle the multi-modal behaviors of SVs, stochastic model predictive control (MPC) approaches have been developed~\cite{jeong2023probabilistic, nair2022stochastic}. For instance, interactive multiple model filters~\cite{jeong2023probabilistic} and Gaussian mixture models~\cite{nair2022stochastic} are employed to estimate the likelihood of the behavior of SVs,  and then a chance-constrained optimization is used to address uncertainties at uncontrolled intersections. Despite their effectiveness, these approaches typically assume that future positions of SVs can be predicted.
In contingency scenarios, such as occluded intersections, predicting the future movement of PVs is challenging. These vehicles may appear suddenly without prior observations, posing a safety risk to the EV~\cite{gilroy2021overcoming}. 

To address contingencies in occluded environments, active perception approaches~\cite{ Zhan-RSS-21, higgins2021negotiating, firoozi2024occlusion,jia2024learning} provide promising solutions. In~\cite{Zhan-RSS-21}, a zero-sum dynamic game is proposed to model potential interactions in dynamic and occluded driving scenarios through game-theoretic active perception. While this method solves a reach-avoid game to avoid collisions with hidden obstacles, it is limited to one-to-one agent interaction scenarios. Such a focus undermines safety in dense traffic environments where {simultaneous interactions with multiple PVs} are prevalent. Additionally, in~\cite{higgins2021negotiating,firoozi2024occlusion}, the EV actively deviates from its reference path to {increase perceptual coverage around occluded regions}, enabling safer navigation. However, this behavior compromises driving consistency, as expert human drivers typically maintain their intended paths in common occluded scenarios, such as intersections~\cite{park2023occlusion}. Other notable occlusion-aware motion planning approaches involve the use of reachable sets~\cite{orzechowski2018tackling, yu2019occlusion, koschi2020set, debada2021occlusion, park2023occlusion, van2024overcoming}. In~\cite{koschi2020set}, an over-approximation approach is developed to approximate the forward reachable set (FRS) of occluded traffic participants for online verification and fail-safe trajectory planning~\cite{pek2018computationally}. While this approach ensures safety, the inherent overestimation of risk typically leads to overly conservative driving behaviors, which could compromise travel efficiency. To address this problem, a probabilistic risk assessment method based on reachability analysis has been introduced in~\cite{yu2020risk}. This method uses a sampling-based technique to model potential {PVs} as particles within the occluded intersection. While effective, this approach is computationally intensive, particularly in dense urban traffic intersections. To facilitate computation, an efficient simplified reachability quantification (SRQ)  has been proposed for occlusion-aware risk assessment to obtain a risk-based speed boundary for the planner \cite{park2023occlusion}.

To enhance safety and driving consistency in unexpected scenarios, contingency planning has emerged as an {effective} approach for rapidly responding to sudden events~\cite{alsterda2019contingency,phiquepal2021control, mustafa2024racp,schweidel2022driver, li2023marc,lyu2024safety, lin2025contingency,packer2023anyone}. Unlike traditional MPC, contingency planning computes two concurrent sequences: a nominal branch, optimized for the intended trajectory, and a contingency branch, which preserves a safe alternative in case of sudden changes~
\cite{alsterda2021contingency}. This dual-branch strategy ensures that a feasible trajectory is always available. To ensure persistent feasibility during planning, terminal set constraints are employed~\cite{schweidel2022driver}. This work incorporates backward reachable sets and maximum control-invariant sets to guarantee that the system can maintain feasible trajectories over the planning horizon. {However, existing contingency planning schemes typically lack explicit occlusion assessment {\cite{li2023marc, phiquepal2021control, mustafa2024racp,lin2025contingency}}. To further account for potential PVs in occluded environments, an {occlusion-aware} contingency game is introduced in a receding horizon planning framework~\cite{qiu2024inferring}. Despite these advancements, scaling contingency planning to handle multi-vehicle interactions in real time remains challenging, particularly in occluded dense traffic environments.} To facilitate computational efficiency, the alternating direction method of multipliers (ADMM)-based optimization offers a promising solution~\cite{taylor2016training, boyd2011distributed}.
By iteratively solving subproblems corresponding to different variables or subsets, ADMM achieves better computational efficiency compared to traditional optimization methods~\cite{peters2024contingency,lyu2019multivehicle}. This property makes ADMM well-suited for real-time applications in occluded or unpredictable scenarios, where rapid response and adaptability are essential.

 In this work, we present an occlusion-aware contingency planner. By coupling online reachability analysis for risk quantification with a consensus ADMM formulation, we decompose the high-dimensional planning problem into a series of low-dimensional convex subproblems. This enables the EV to regenerate exploration and fallback trajectories in real time, preserving scalability with respect to the number of SVs in dynamic, occluded environments. The main contributions of this paper are summarized as follows: 
\begin{itemize}
	\item We introduce an occlusion-aware contingency safety-critical planning approach that simultaneously optimizes two trajectories: one that prioritizes exploration to maximize situational awareness, and the other that serves as a safety fallback to enhance safe navigation for potential hazards based on reachability analysis. Both trajectories share an initial segment, allowing smooth transitions to handle emerging or occluded PVs. 

	\item We leverage the consensus ADMM to facilitate real-time optimization in our dual-trajectory planning framework. The consensus ADMM approach decomposes the high-dimensional planning problem into smaller subproblems that can be solved in an alternating optimization manner. This enables rapid computation of exploration and fallback trajectories within a receding horizon planning framework, ensuring real-time adaptation to dynamic occluded environments.

	\item We validate our occlusion-aware contingency planning algorithm through simulations and real-world experiments using a 1:10 scale Ackermann mobile robot platform. The results demonstrate the effectiveness of the proposed approach in generating safe and efficient trajectories for AVs in occluded intersections. Additionally, a detailed evaluation of computational efficiency demonstrates the algorithm's feasibility for real-time application in dynamic and traffic-heavy environments with varying obstacle conditions.  
\end{itemize}

The rest of the paper is structured as follows: 
Section~\ref{sec:ProblemStatement} introduces the problem formulation and preliminaries, including spatiotemporal barriers and trajectory parameterization. Section~\ref{sec:Methodology} presents the proposed occlusion-aware contingency planning approach and provides a detailed problem decomposition. Section~\ref{sec:Results} demonstrates the effectiveness of the proposed approach through both simulation and real-world experiments. Finally, Section~\ref{sec:Conclusions} concludes the paper and discusses the future work.
 
\section{Problem Formulation and Preliminaries}\label{sec:ProblemStatement}

\begin{figure}[tb]
\begin{center}
\includegraphics[width=.95\columnwidth]{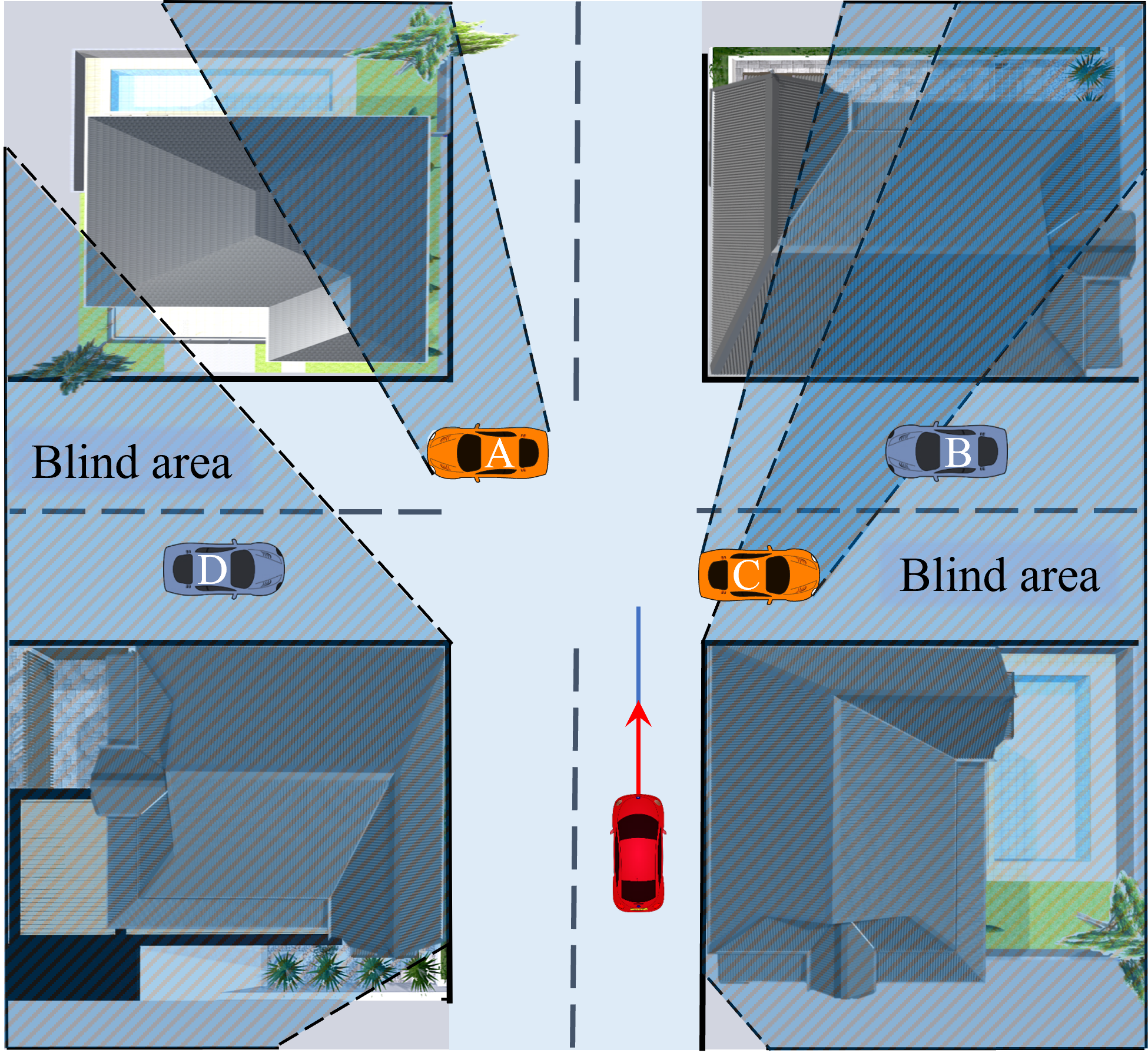}    \vspace{-1mm}
\caption{ The EV (red) navigates an occluded four-way intersection, generating two trajectories: a blue exploration trajectory and a red fallback trajectory with an arrow. Both trajectories share an initial common segment to enable smooth transitions between trajectories. Visibility is obstructed by static obstacles (buildings) and dynamic vehicles (orange), highlighting the challenges of safe and efficient driving in occluded environments.  
} \vspace{-4mm}
\label{fig:occ}
\end{center}
\end{figure} 
\subsection{Problem Formulation} This study addresses the challenge of enabling safe and efficient navigation for the EV in occluded areas, where visibility is hindered by static obstacles and dynamic SVs, as illustrated in Fig.~\ref{fig:occ}. 
The dynamics of the EV are modeled using a Dubin's car model~\cite{chen2024interactive}: 
\begin{equation}
 {\textbf{x}}=\left[\begin{array}{c}
 {p}_x\\
 {p}_y\\
 {\theta}\\
 {v} 
\end{array}\right], \textbf{u}=\left[\begin{array}{c}
{u_{\theta}} \\ 
{a} 
\end{array}\right] 
 ,\textbf{x}^{+} = \left[\begin{array}{c}
 {p}_x + v\cos \left( \theta\right) \Delta t\\
 {p}_y  +v\sin \left(\theta \right) \Delta t\\
 {\theta} +  {u_{\theta}} \Delta t\\ 
v + {a} \Delta t
\end{array}\right] ,
\label{eq:ev_model}
\end{equation} 
where {${\textbf{x}}$, ${\textbf{u}}$, and ${\textbf{x}}^{+}$ denote the current state vector, the control input vector, and the state vector at the next time step, respectively;} ${p}_x$ and ${p}_y$ denote the longitudinal and lateral {positions}, respectively; ${\theta}$ and $v$ denote the heading angle and velocity, respectively;  $\Delta t$ denotes the discrete time step, and the control inputs are the yaw rate \(u_{\theta}\) and acceleration \(a\).  To simplify the problem, the following assumptions are made:
\begin{assumption}(\textbf{Perception Capabilities}~\cite{van2024overcoming})\label{assumption: communication}
The EV is equipped with perceptual sensors that can accurately detect free space within its field of view.
\end{assumption}   
\begin{assumption}({\textbf{Velocity Distribution}~\cite{yu2019occlusion,park2023occlusion}})\label{assumption:velocity}  
The velocity of the {PV} is assumed to remain constant and is uniformly distributed within the range $[0, v_{\text{pv},\max}]$, where $v_{\text{pv},\max}$ denotes the maximum possible velocity of the PV.  
\end{assumption}

To address the challenges of safe navigation in occluded areas, we introduce the concepts of Forward Reachable Set (FRS) and Backward Reachable Set (BRS): \begin{definition}[\textbf{Reachable Sets: Forward and Backward}]
Let \( \mathcal{X} \subset \mathbb{R}^n \) denote the state space and \( \mathcal{U} \subset \mathbb{R}^m \) denote the control input space. The system dynamics are described by \( f: \mathcal{X} \times \mathcal{U} \to \mathcal{X} \), where \( \dot{\mathbf{x}}(t) = f(\mathbf{x}(t), \mathbf{u}(t)) \).  Let \( \phi(t, \mathbf{x}_0, \mathbf{u}) \) denote the state trajectory of the system, starting from \( \mathbf{x}_0 \in \mathcal{X} \) at \( t = 0 \), under the control input \( \mathbf{u}(\cdot): [0, T] \to \mathcal{U} \). 

1. \textbf{Forward Reachable Set (FRS):} The FRS over a time horizon \( T > 0 \), starting from an initial set \( \mathcal{X}_0 \subset \mathcal{X} \), is defined as the set of all states reachable within time \( T \) under admissible controls, as follows:
\[
\begin{aligned}
\operatorname{FRS}(T,\mathcal X_0)
=\big\{\mathbf x\in\mathcal X\mid\;&
\exists\,\mathbf x_0\in\mathcal X_0,\;
\exists\,\mathbf u(\cdot):[0,T]\to\mathcal U,\\
&\exists\,t\in[0,T]:
\mathbf x=\phi(t,\mathbf x_0,\mathbf u(\cdot))
\big\}.
\end{aligned}
\]

2. \textbf{Backward Reachable Set (BRS):}  The BRS over a time horizon \( T > 0 \), for a target set \( \mathcal{T} \subset \mathcal{X} \), is the set of all initial states from which the trajectory can reach set \( \mathcal{T} \) at some time \( t \in [0, T]\) under admissible controls, as follows:
\[
\begin{aligned}
\text{BRS}(T,\mathcal T)
=\big\{\mathbf x_0\in\mathcal X\mid\;&
\exists\,\mathbf u(\cdot):[0,T]\to\mathcal U,\\
&\exists\,t\in[0,T]:
\phi(t,\mathbf x_0,\mathbf u(\cdot))\in\mathcal T
\big\}.
\end{aligned}
\]

\end{definition} 
As outlined in~\cite{park2023occlusion}, the set $\mathcal{T}$ of a PV is typically treated as the “avoid set” to ensure safe driving, which is computed using the FRS. A general contingency motion planning problem can then be formulated as follows:
\begin{subequations}\label{eq:problem1}
    \label{problem}\begin{align}
    \displaystyle\operatorname*{minimize}_{\mathbf{x}, \mathbf{u}}~~~
    & \mathcal{L}\label{eq:problem}\\
    \quad\text{subject to}\quad  &\mathbf{x}^{\text{e}}_{0}=\mathbf{x}^{\text{s}}_{0}=\mathbf{x}(t_0), \label{eq:problem_1}\\
    &\mathbf{x}^{\text{e}}_{N} \in \mathcal{X}^{\text{e}}_f, \quad \mathbf{x}^{\text{s}}_{N} \in \mathcal{X}^{\text{s}}_f, \label{eq:problem_2}\\
    & \mathbf{x}_{k+1}  = f^{EV}(\mathbf{x}_{k}, \mathbf{u}_{k}),  \label{eq:problem_3}\\ 
    & \big(\mathbb{X}^{\text{e}}_k \cup \mathbb{X}^{\text{s}}_k \big)\cap \mathbb{O}^{(i)}_k = \emptyset,    \quad  \label{eq:problem_4}\\
    & \mathbf{x}^{\text{e}}_{k} \in  \mathcal{X}^{\text{e}}, \quad \mathbf{x}^{\text{s}}_{k} \in  \mathcal{X}^{\text{s}},   \label{eq:problem_7}\\ 
    & \mathbf{u}_{k} \in  \mathcal{U},   \label{eq:problem_6}\\ 
     & \varphi( \mathbf{x}^{\text{e}}_{l}) =\varphi( \mathbf{x}^{\text{s}}_{l}) = \varphi(\Tilde{\mathbf{x}}_{l}), \label{eq:problem_5}\\ 
    & \forall k\in\mathcal{I}_0^{N-1}, i\in\mathcal{I}_0^{m-1},  l\in\mathcal{I}_0^{N_s-1}. \nonumber \vspace{-2mm}
\end{align}
\end{subequations} 
Here, \(\mathbf{x}^{\text{e}}_k\) and \(\mathbf{x}^{\text{s}}_k\) denote the state vectors for the exploration and safety fallback trajectories at time step \(k\), respectively; and \(\mathbf{u}_k\) is the control input. These state vectors evolve according to the vehicle dynamics \(f^{EV}\) as defined in \eqref{eq:ev_model}. The initial state constraint \eqref{eq:problem_1} ensures both trajectories start from the same state, while the terminal constraint \eqref{eq:problem_2} ensures they reach their respective goal sets. Note that the state spaces \(\mathcal{X}^{\text{e}}\) and \(\mathcal{X}^{\text{s}}\) are dynamically refined using reachability analysis to account for environmental changes \cite{park2023occlusion}.  Building on this, a spatiotemporal barrier is employed to enforce safety constraints. The safety constraint \eqref{eq:problem_4} ensures that the EV's reachable sets do not overlap with the obstacles’ occupancy sets. It is pertinent to note that:
\begin{itemize}
    \item \(\mathbb{X}^{\text{e}}_k\) and \(\mathbb{X}^{\text{s}}_k\) are the compact reachable sets of the EV along the exploration and fallback trajectories, respectively, at time step \(k\).
    \item \(\mathbb{O}^{(i)}_k\) represents the compact occupancy set of the \(i\)-th surrounding obstacle or vehicle at time \(k\).
\end{itemize} 
\begin{remark} SVs are excluded from these safety constraints if their FRSs do not intersect with the EV's FRS. For instance, as illustrated in Fig.~\ref{fig:occ}, vehicles A and C are excluded from consideration since their FRSs do not overlap with the EV's FRS. \end{remark}

To maintain the trajectory consistency, a global consistency variable \(\Tilde{\mathbf{x}}_{l}\) is introduced, and the function $\varphi(\cdot)$ extracts the desired variable from the states. The consistency constraint \eqref{eq:problem_5} ensures that the exploration and fallback trajectories share a common segment of length \(N_s\).

The objective function $ \mathcal{L}$ minimizes the control effort and goal-tracking errors, defined as follows:  
\begin{alignat}{2}
    \small    
    \mathcal{L}  = & \sum_{k=0}^{N-1} \|\mathbf{u}^{\text{e}}_k \|^2_2 + {w_{1}} \|\mathbf{u}^{\text{s}}_k \|^2_2 + \sum_{k=N_s + N_d}^{N-1} {w_2}\| \mathbf{x}^{\text{e}}_k -   \mathbf{x}^{\text{e}}_{goal}\|^2_2 \notag   \\  \vspace{-2mm}
   & +  \sum_{k=N_s + N_d}^{N-1}  {w_3}\| \mathbf{x}^{\text{s}}_k -   \mathbf{x}^{\text{s}}_{goal}\|^2_2,
    \label{eq:obj_func} \vspace{-2mm}
\end{alignat}
where \(w_1\), \(w_2\), and \(w_3\) are constant positive weights; $N$ denotes the horizon of the trajectory;   $N_d$ denotes the free steps after two trajectories diverge.  

With this problem formulation, two main challenges arise:
\begin{itemize} 
\item Balancing driving safety and efficiency to avoid overly conservative or aggressive driving behaviors. 
\item Efficiently solving this nonlinear optimization problem while adhering to the non-convex safety constraint~\eqref{eq:problem_4} and kinematic constraint~\eqref{eq:problem_3}. 
\end{itemize} 

 In this study, we adopt a trajectory selection mechanism that considers multiple factors, including goal-tracking, lateral deviation, safety, comfort, and consistency costs, as detailed in our prior work~\cite{zheng2025barrier}. Notably, both the exploration and fallback trajectories share a common initial segment. This structure ensures that the next immediate control action executed by the EV is identical for both trajectories, regardless of the final trajectory chosen, thereby facilitating smooth and stable transitions within a receding horizon planning framework. 

\subsection{Spatiotemporal Barrier}
In this work, we employ the following spatiotemporal barrier safety constraint, as outlined in~\cite{rastgar2020novel,zheng2025barrier}, to enforce the safety constraint \eqref{eq:problem_4}:
\begin{eqnarray}
\left\{
    \begin{aligned}
      p_{x,k} &= o_{x,k}^{(i)} + l_x^{(i)}   \xi ^{(i)}_k  \cos (\omega^{(i)}_k), \\
      p_{y,k} &= o^{(i)}_{y,k} + l_y^{(i)}   \xi ^{(i)}_k \sin (\omega^{(i)}_k),  \\
   \Delta  h&{(\mathbf{x}_k, \mathbf{o}^{(i)}_k) + \alpha_k  h(\mathbf{x}_{k-1}, \mathbf{o}^{(i)}_{k-1})  > 0} ,   \forall  i\in\mathcal{I}_0^{M-1}.   
    \end{aligned}
\right.
\label{eq:polar_safety}
\end{eqnarray} 
Here, \( p_{x,k} \) and \( p_{y,k} \) represent the longitudinal and lateral positions of the EV at time step \(k\), while \( o_{x,k} \) and \( o_{y,k} \) are the longitudinal and lateral positions of the \(i\)-th obstacle; $\mathbf{o}^{(i)}_k =  [o_{x,k}^{(i)} \quad o_{y,k}^{(i)} \quad o_{v_x,k}^{(i)} \quad o_{v_y,k}^{(i)}]^T $ denotes the position and velocity of the $i$-th obstacle at time step $k$. The terms \( l_x^{(i)} \) and \( l_y^{(i)} \) define the axes of the safety ellipse for the \(i\)-th obstacle, with \(  \xi _k^{(i)} >  1 \) ensuring a safety margin between the EV and the \(i\)-th obstacle at time step \(k\). The barrier coefficient $\alpha_k$ balances the barrier's influence over successive time steps. The barrier function \( h \) is defined as 
\begin{equation}
    h(\mathbf{x}_k, \mathbf{o}^{(i)}_k) =  \xi^{(i)}_k - 1,
    \label{eq:barrier_func}
\end{equation} 

The angle \( \omega^{(i)}_k \) at time step \(k\) between the EV and the \(i\)-th obstacle is given by 
\begin{equation}
    \omega^{(i)}_k = \arctan\left( \frac{l_{x,k}^{(i)}( p_{y,k}  - o_{y,k}^{(i)})}{l_{y,k}^{(i)}( p_{x,k}  - o_{x,k}^{(i)})} \right).\vspace{-2mm}
    \label{eq:omega}
\end{equation}

\subsection{Trajectory Parameterization}
\label{subsec:Trajectory_para}   
We parameterize the EV's trajectory using \( m \)-dimensional, \( n \)-th order B\'ezier curves. The trajectory for each curve is defined as  
\begin{equation}
    \mathbf{C}^{(j)}(\nu) = \sum_{i=0}^{n}  {B}_{i,n} (\nu) \mathbf{P}^{(j)}_{i}, \quad j \in \mathcal{I}_0^{N_c-1},\vspace{-2mm}
\end{equation}
where \( N_c \) represents the total number of trajectories; \( \mathbf{P}^{(j)}_{i} \in \mathbb{R}^m \) denotes the control points or B\'ezier coefficients that are optimized for the \( j \)-th trajectory. For the EV, these control points are defined as \( \mathbf{P}^{(j)}_{i} = [c^{(j)}_{x,i} \quad c^{(j)}_{y,i} \quad c^{(j)}_{\theta,i}]^T \), where \( c^{(j)}_{x,i} \), \( c^{(j)}_{y,i} \), and \( c^{(j)}_{\theta,i} \) represent the longitudinal position coefficient, lateral position coefficient, and heading angle coefficient, respectively. The Bernstein polynomial basis \( {B}_{i,n} \) is defined as follows: 
\[
{B}_{i,n} (\nu) = \binom{n}{i} \nu^i (1 - \nu)^{n - i},
\]
with \( \nu = \frac{t - t_0}{T} \in [0, 1] \), where \( t_0 \) is the initial time. The variable \( t = t_0 + k\delta t \) indicates the current time in the trajectory, with \( k \) representing discrete time steps and \( \delta t = T/N \) as the time interval.

Consequently, the trajectory can be expressed in a discrete form, as follows:
\[
\left\{ \mathbf{C}^{(j)}_k \right\}_{k=0}^{N-1} = \mathbf{W}^T_{P,j} \mathbf{W}_B, \quad j \in \mathcal{I}_0^{N_c-1},
\]
where \( \mathbf{C}^{(j)}_k = [p^{(j)}_{x,k} \quad p^{(j)}_{y,k} \quad \theta^{(j)}_{k}]^T \) denotes the trajectory's longitudinal position, lateral position, and heading angle at time step \( k \). The control points matrix \( \mathbf{W}_{P,j} \) and the constant basis functions matrix \( \mathbf{W}_B \) are defined as follows:
 \begin{align}
     \mathbf{W}_{P,j}&= [\mathbf{c}_x^{(j)}\quad \mathbf{c}_y^{(j)}\quad \mathbf{c}_{\theta}^{(j)}] \notag\\&= [\mathbf{P}^{(j)}_0 \quad \mathbf{P}^{(j)}_1 \quad \dots \quad \mathbf{P}^{(j)}_n]^T \in \mathbb{R}^{(n+1)\times 3},
 \end{align}  
\begin{equation}
\mathbf{W}_B = [\mathbf{B}_{0} \quad \mathbf{B}_{1} \quad \dots \quad  \mathbf{B}_{n}]^T \in \mathbb{R}^{(n+1)\times N}. \end{equation} 
In this study, we assign $j=0$ and $j=1$ to denote the exploration and fallback trajectories, respectively. Therefore, the goal is to optimize the control point matrices \( \mathbf{C}_x = \left[ \mathbf{c}_x^{(0)} \quad \mathbf{c}_x^{(1)} \right] \in \mathbb{R}^{(n+1)\times 2} \), \( \mathbf{C}_y = \left[ \mathbf{c}_y^{(0)} \quad \mathbf{c}_y^{(1)} \right] \in \mathbb{R}^{(n+1)\times 2} \), and \( \mathbf{C}_\theta = \left[ \mathbf{c}_\theta^{(0)} \quad \mathbf{c}_\theta^{(1)} \right] \in \mathbb{R}^{(n+1)\times 2} \).
 
In this study, the order $n$  is set to be 10. Notably, 
{these high-order B\'ezier polynomials inherently impose motion constraints on the EV, ensuring alignment with the kinematics of discrete integrator models} and facilitating smooth transitions in position, velocity, and acceleration. 

\section{Methodology}\label{sec:Methodology}
 In this section, we present our contingency motion planning approach in occluded environments. First, we employ reachability analysis to assess risks arising from occlusions. Based on this analysis, we quantify the risk-aware state spaces \(\mathcal{X}^{\text{e}}\) and \(\mathcal{X}^{\text{s}}\), corresponding to the exploration trajectory and safety fallback trajectory, respectively. {The contingency motion planning problem~\eqref{problem} is then reformulated as a biconvex optimization problem that explicitly incorporates these risk considerations.} Finally, we leverage the consensus ADMM to decompose and efficiently solve this reformulated problem.  { The overall computational workflow of the contingency planner is summarized in Algorithm~\ref{alg:occlusion_planner}.} 

\subsection{Occlusion-Aware Risk Assessment} 
Navigating occluded scenarios, such as intersections or roads with parked vehicles, presents critical challenges for autonomous driving systems. In these contexts, expert human drivers typically adjust their speed rather than alter their path to mitigate risks. {We employ the SRQ method~\cite{park2023occlusion} to assess risks and determine safe velocity boundaries for  planning.}

\subsubsection{Risk Quantification Through SRQ} 
 {The SRQ method quantifies risk within a dynamic \textit{Phantom Vehicle Set (PVS)}, which represents the occluded zone where {PVs} might exist.  
The calculation is based on two core assumptions: (1) PVs drive along the lane centerline, and (2) with no prior information, a PV’s initial position is uniformly distributed across the PVS, defined by its start and end points \(s_s\) and \(s_e\).} 

 Hence, we can calculate the number of potential PVs, $g(s)$, at the final relevant risk position $s$ according to the $BRS(T, s)$, which represents all positions a PV could occupy within the prediction horizon \( T \). The domain of \(g(s)\) is divided into three intervals: $ 
I_1 := [s_s, s_e],\  
I_2 := [s_e, s_s + v_{\text{pv},\text{max}} T],\  
I_3 := [s_s + v_{\text{pv},\text{max}} T, s_e + v_{\text{pv},\text{max}} T].
$  The function \(g(s)\) is defined piecewise as:
\begin{equation}
g(s) = 
\begin{cases} 
\frac{1}{2} \left( 2v_{\text{pv},\text{max}} - \frac{s - s_s}{T} \right)(s - s_s), & s \in I_1 ,\\ 
\frac{1}{2} \left( 2v_{\text{pv},\text{max}} - \frac{s - s_s}{T} - \frac{s - s_e}{T} \right)(s_e - s_s), & s \in I_2, \\ 
\frac{1}{2} \left( v_{\text{pv},\text{max}} - \frac{s - s_e}{T} \right)\left(s_e - (s - v_{\text{pv},\text{max}} T)\right), & s \in I_3.
\end{cases}
\end{equation}

The longitudinal risk \( r_{\text{lon}}(s) \) is then obtained by scaling \(g(s)\) by the length of the PVS, amplifying the risk for larger occluded zones:
\begin{equation}
r_{\text{lon}}(s) = (s_e - s_s) \cdot g(s).
\end{equation}

To account for uncertainties in the lateral position of PVs, lateral deviations $d$ are modeled as a normal distribution:
\begin{equation}
r_{\text{lat}}(d)
=
\mathcal{N}\left(
d;\,0,\,
\left(\frac{l_w}{2Z(1-0.5(1-\eta))}\right)^2
\right),
\end{equation}
where $l_w$ is the lane width, and
$Z(p)=\Phi^{-1}(p)$ is the standard normal quantile function.  A deviation of \( d = 0 \) corresponds to PVs positioned at the center of the lane, where the likelihood (i.e., risk) is the highest. The total risk is computed by combining the longitudinal and lateral risks, as follows:
\begin{equation} 
\end{equation}

\begin{figure}[tb]
\begin{center}
\includegraphics[width=.98\columnwidth]{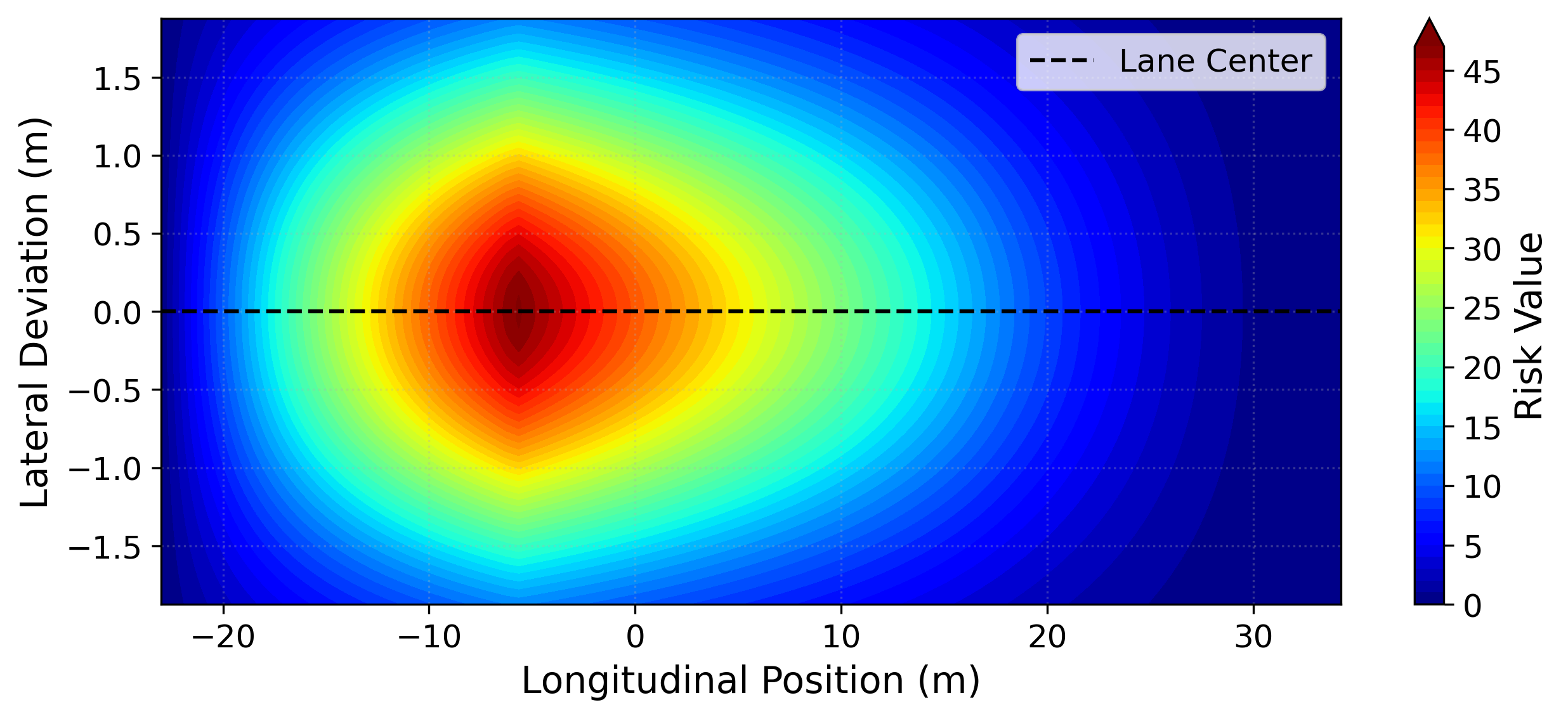}     \vspace{-2mm}
\caption{Spatial distribution of risk around the EV in the presence of occlusion. The EV is positioned at [$0\,\text{m}$, $-2\,\text{m}$], with the occluded area extending from [$-8\,\text{m}$, $0\,\text{m}$] to [$-2\,\text{m}$, $0\,\text{m}$].  {The quantitative color scale (right), labeled `Risk Value', illustrates the risk level, where warmer colors (e.g., red) represent higher risk and cooler colors (e.g., blue) indicate lower risk.}} 
\label{fig:risk_map} \vspace{-6mm}
\end{center}
\end{figure}

In this study, we assess the risk around the EV by considering parameters: a maximum PV velocity of $10\,\text{m/s}$, a prediction horizon of $4 \,\text{s}$, and a lane width of $3.75 \,\text{m}$. We use a 90\% coverage probability
($\eta=0.90$, $Z(0.95)\approx1.645$) for lateral deviation, which corresponds to $\pm 1.645$ standard deviations ($\sigma$). The spatial risk distribution is shown in Fig.~\ref{fig:risk_map}.  

\subsubsection{Dynamic Velocity Sets} 
\label{subsubsec:dynamic_vel}
Once the risks $r(s, d)$ are quantified, they are used to dynamically adjust the maximum velocity boundary $v_{\text{occ,s}}$, as follows: 
\begin{equation}
    v_{\text{occ,s}} = 
    \begin{cases} 
        v_{\text{occ,min}}, & \text{if } r_{\text{total}}  > c_{\text{th,max}}, \\
        \Delta v (r_{\text{total}} - c_{\text{th,min}}) + v_{\text{occ,max}}, & \text{otherwise}.
    \end{cases}
    \label{eq:occ_vel}
\end{equation}
Here, $c_{\text{th,min}}$ and $c_{\text{th,max}}$ denote predefined minimum and maximum risk thresholds, respectively; $\Delta v$ denotes the change rate of the velocity, defined as follows:
\begin{equation}
    \Delta v = \frac{v_{\text{occ,min}} - v_{\text{occ,max}}}{c_{\text{th,max}} - c_{\text{th,min}}},
\end{equation} 
where $v_{\text{occ,min}}$ and $v_{\text{occ,max}}$ denote the minimum and maximum allowable speeds of the EV under occlusion, and $r_{\text{total}}$ denotes the aggregated risk for occluded zones.
{To strike a balance between travel efficiency and safety in occluded areas, two distinct risk thresholds are employed, yielding two dynamic maximum velocity boundaries: \( v^0_{\text{occ,s}} \) for the exploration trajectory and \( v^1_{\text{occ,s}} \) for the fallback trajectory.  These thresholds ensure that exploration trajectories prioritize situational awareness while fallback trajectories maintain a safe alternative. }
\begin{remark}
   As outlined in \cite{park2023occlusion}, occlusion risks \(r(s, d)\) can be ignored when the FRS of the PVs does not intersect with the planned trajectory of the EV within its prediction horizon. 
\end{remark} 

\subsection{Consensus ADMM for Contingency Motion Planning} 
\subsubsection{Reformulation of Contingency Motion Planning}  
We reformulate the objective function \eqref{eq:obj_func} to ensure that the EV remains in its target lane and achieves the desired longitudinal velocity \( v_{x,d} \),  as follows:
\begin{alignat}{2}
    \small    
      \mathcal{L}  =  & \frac{1}{2} \mathbf C_\theta^T\mathbf Q_\theta\mathbf C_\theta 
+\frac{1}{2}\mathbf C_x^T\mathbf Q_x\mathbf C_x
+\frac{1}{2} \mathbf C_y^T\mathbf Q_y\mathbf C_y\notag\\
    &+ \frac{1}{2}\|  \dot{\mathbf{W}}^T_B \mathbf{C}_x - \mathbf{V}_{x,d}\|^2_{\mathbf{Q}_{1}} + \frac{1}{2}\|  \mathbf{A}^T_{\text{y},\text{d}} \mathbf{C}_y - \mathbf{P}_{y,d}\|^2_{\mathbf{Q}_{2}} \notag\\
    =  &f(\mathbf{C}_\theta) + g(\mathbf{C}_x) + g(\mathbf{C}_y), 
    \label{eq:obj_func_new} 
\end{alignat} 
where \begin{subequations}\small
\begin{align}  
      &  f(\mathbf{C}_\theta) =  \frac{1}{2} \mathbf{C}^T_\theta\mathbf{Q}_{\theta} \mathbf{C}_\theta, \label{eq:obj_func_new_theta}\ \\
      & g(\mathbf{C}_x) =  \frac{1}{2} \mathbf{C}^T_x\mathbf{Q}_{x} \mathbf{C}_x+ \frac{1}{2}  \mathbf{C}^T_x\dot{\mathbf{W}}_B  \mathbf{Q}_1 \dot{\mathbf{W}}_B^T \mathbf{C}_x -   \mathbf{V}_{x,d}^T \mathbf{Q}_1 \dot{\mathbf{W}}_B^T \mathbf{C}_x   , \label{eq:obj_func_new_x}\\
      &  g(\mathbf{C}_y) =  \frac{1}{2} \mathbf{C}^T_y\mathbf{Q}_{y} \mathbf{C}_y+ \frac{1}{2} 
 \mathbf{C}^T_y\mathbf{A}_{\text{y},\text{d}} \mathbf{Q}_2 \mathbf{A}^T_{\text{y},\text{d}} \mathbf{C}_y -  \mathbf{P}_{y,d}^T \mathbf{Q}_2 \mathbf{A}^T_{\text{y},\text{d}} \mathbf{C}_y.\label{eq:obj_func_new_y}\
   \end{align}      
\end{subequations} 
Here, $\mathbf{Q}_{\theta}$, $\mathbf{Q}_{x}$, $\mathbf{Q}_{y}$,  $\mathbf{Q}_{1}$, and $\mathbf{Q}_{2}$ are symmetric positive semi-definite matrices; \( \mathbf{V}_{x,d} \in \mathbb{R}^N \) is a vector in which each element is equal to \( v_{x,d} \); \( \mathbf{P}_{y,d} \in \mathbb{R}^{N - N_s - N_d} \) is the lateral position vector along the current lane centerline, with each element corresponding to the lateral position $p_{y,d}$. The matrix \( \mathbf{A}_{y,d} \in \mathbb{R}^{(n+1) \times (N - N_s - N_d)} \) maps the lateral control points of trajectories to the centerline of the driving lane, defined as 
\[
\mathbf{A}_{y,d} = [  
    \mathbf{B}_{0,[N_s + N_d : N]} \ \mathbf{B}_{1,[N_s + N_d : N]} \ \dots \ \mathbf{B}_{n,[N_s + N_d : N]} ]^T.
\]

We reformulate the initial contingency motion planning problem \eqref{eq:problem1} as a biconvex NLP problem by incorporating the spatiotemporal control constraints \eqref{eq:polar_safety} and dynamic velocity boundaries from Section~\ref{subsubsec:dynamic_vel}, as follows: \begin{subequations}
\label{eq1:problem}
\begin{align} 
 \hspace{-4mm}   \displaystyle\operatorname*{min}_{\substack{
 \{\mathbf{C}_x, \mathbf{C}_y, \mathbf{C}_\theta; \\
        \quad\quad\quad \bm{\omega}, \bm{\xi}\} 
      }}~~ 
    \ &\ \mathcal{L}
    \label{eq1:problem0} \\
    \quad\text{s.t.}\quad
    & \mathbf{A}_0 [  \mathbf{C}_x \quad  \mathbf{C}_y\quad \mathbf{C}_{\theta} ]  = [\mathbf{E}_{x,0}\quad  \mathbf{E}_{y,0}\quad \mathbf{E}_{\theta,0}], \label{eq1:problem_1} \\
    & \mathbf{A}_{f,y} \mathbf{C}_y   = \mathbf{E}_{y,f},\quad   \mathbf{A}_{f,\theta}\mathbf{C}_{\theta}   = \mathbf{E}_{\theta,f} , \label{eq1:problem_2}     \end{align}
     \begin{align}
   &   \dot{\mathbf{W}}^T_B  \begin{bmatrix} 
 \mathbf{C}_x \\   \mathbf{C}_y  \end{bmatrix}
    = \mathbf{V} \cdot 
    \begin{bmatrix}
        \cos{ (\mathbf{W}^T_B \mathbf{C}_\theta)} \\
        \sin{ (\mathbf{W}^T_B \mathbf{C}_\theta)}
    \end{bmatrix} , \label{eq1:problem3}\\    
    & \mathbf{A}_h \mathbf{C}_{x}  = \mathbf{O}_{x}  + \mathbf{L}_x \cdot  \bm{\xi}    \cdot  \cos ( \bm{\omega}), \label{eq1:problem4} \\  
    & \mathbf{A}_h \mathbf{C}_{y} = \mathbf{O}_{y} +\mathbf{L}_y \cdot  \bm{\xi} \cdot\sin ( \bm{\omega}),\label{eq1:problem5} \\
   & \bm{\omega}  \in \mathcal{C}_{\omega}, \bm{\xi}  \in \mathcal{C}_{\xi}, \label{eq1:problem6}\\ 
     &  \begin{bmatrix}
         \dot{\mathbf{W}}^T_B \\
        - \dot{\mathbf{W}}^T_B
    \end{bmatrix}  \mathbf{C}_x \leq  \begin{bmatrix}
         \mathbf{V}_{\text{occ},\text{max}}  \\
        - \mathbf{V}_{\text{occ},\text{min}} 
    \end{bmatrix}    ,\label{eq1:problem7}\\
    & \mathbf{G}  \mathbf{C}_x  \leq \mathbf{F}_x ,\label{eq1:problem8}\\
    & \mathbf{G}  \mathbf{C}_y \leq \mathbf{F}_y ,\label{eq1:problem9} \\
      & \mathbf{A}^T_{\text{c},x} \mathbf{C}_x   = \mathbf{Z}_{x}, 
       \mathbf{A}^T_{\text{c},y}  \mathbf{C}_y    = \mathbf{Z}_{y},\label{eq1:problem10}  \\ 
      &  \mathbf{A}^T_{\text{c},\theta}  \mathbf{C}_{\theta}     = \mathbf{Z}_{\theta} .\label{eq1:problem11}  
 \end{align}
\end{subequations}
Notably, the constraint \eqref{eq1:problem_1} enforces initial conditions. The vectors \( \mathbf{E}_{x,0}, \mathbf{E}_{y,0}, \mathbf{E}_{\theta,0} \in \mathbb{R}^4 \) represent the initial position and velocity constraints in the longitudinal, lateral, and orientation components, respectively, in both exploration and fallback trajectories. The matrix $\mathbf{A}_0$ is defined as follows:
\begin{equation} \vspace{-0mm}
\mathbf{A}_0  =  \begin{bmatrix}  
    \mathbf{B}_{0,1}\quad \mathbf{B}_{1,1}\quad  \dots\quad \mathbf{B}_{n,1} \\
    \dot{\mathbf{B}}_{0,1} \quad\dot{\mathbf{B}}_{1,1} \quad \dots\quad \dot{\mathbf{B}}_{n,1} \end{bmatrix},
\notag
\vspace{-0mm}
\end{equation} 
where \( \dot{\mathbf{B}}_{i,j} \) denotes the time derivative of the Bernstein polynomial basis matrix \( \mathbf{B}_{i,j} \).

The terminal constraint \eqref{eq1:problem_2} ensures that the final lateral positions of the trajectories align with the centerline while maintaining a small heading angle for enhanced driving stability. The associated matrices are given by   \begin{equation} \vspace{-0mm} 
 \mathbf{A}_{f,y}  =  [\mathbf{B}_{0,N}\quad \mathbf{B}_{1,N}\quad  \dots\quad \mathbf{B}_{n,N}   ] , \mathbf{E}_{f,y} = [p_{y,d}\quad p_{y,d}]^T,
\notag
 \vspace{-0mm} \end{equation}  
 \[ \vspace{-0mm}\mathbf{A}_{f,\theta} = \begin{bmatrix}\mathbf{B}_{0,N}\quad \mathbf{B}_{1,N}\quad  \dots\quad \mathbf{B}_{n,N}   \\
\dot{\mathbf{B}}_{0,N} \quad\dot{\mathbf{B}}_{1,N} \quad \dots\quad \dot{\mathbf{B}}_{n,N} \end{bmatrix} , \mathbf{E}_f=\mathbf{0} \in \mathbb{R}^{4}.
\notag
  \vspace{0mm} \]
 
As outlined in~\cite{zheng2025barrier}, the nonholonomic motion constraints and collision avoidance constraints are enforced by \eqref{eq1:problem3} and \eqref{eq1:problem4}-\eqref{eq1:problem6}, respectively. In these equations, 
the velocity matrix \(\mathbf{V} = [\mathbf{V}_0\quad \mathbf{V}_1 ] \in \mathbb{R}^{N\times 2}\); the set \( \mathcal{C}_{\xi} \) represents the safety barrier set for \( \bm{\xi} \in \mathbb{R}^{(N \times M) \times 2} \), which is derived from the equations \eqref{eq:polar_safety} and \eqref{eq:barrier_func}. Additionally, the set \( \bm{\omega} \in \mathbb{R}^{(N \times M) \times 2} \) is computed using \eqref{eq:omega} over \( N \) planning steps for \( M \) obstacles in a planning horizon. 
The matrix $\mathbf{A}_h$ is defined as
\[
\mathbf{A}_h = \left[ \mathbf{W}_B^T \otimes \mathbf{I}_M \quad \mathbf{W}_B^T \otimes \mathbf{I}_M \right]^T,
\]
 where \( \mathbf{W}_B^T \otimes \mathbf{I}_M \in \mathbb{R}^{(N \times M) \times (n+1)} \) represents the repeated vertical stacking of the basis function matrix \( \mathbf{W}_B^T \) for \( N \) steps and \( M \) obstacles.  \( \mathbf{L}_x \in \mathbb{R}^{(N \times M) \times 2} \) and \( \mathbf{L}_y \in \mathbb{R}^{(N \times M) \times 2} \) represent the corresponding safety ellipse matrices for \( M \) obstacles along the planning steps \( N \). A similar definition applies to $\mathbf{O}_x\in \mathbb{R}^{(N \times M) \times 2}$ and  $\mathbf{O}_y \in \mathbb{R}^{(N \times M) \times 2}$, which denotes the predicted future longitudinal position and lateral position of \( M \) obstacles over  \( N \) planning steps.

The constraint \eqref{eq1:problem7} enforces dynamic velocity boundaries for occlusion-aware driving tasks, where \( \mathbf{V}_{\text{occ},\text{min}} \in \mathbb{R}^{N \times 2} \) represents the minimum velocity matrix, with each element corresponding to a minimum velocity \( v_{\text{occ},\text{min}} \). The maximum velocity limits matrix $\mathbf{V}_{\text{occ},\text{max}}$ is defined as
\[\mathbf{V}_{\text{occ},\text{max}} = [\mathbf{V}^0_{\text{occ},\text{max}} \quad \mathbf{V}^1_{\text{occ},\text{max}}] \in \mathbb{R}^{N \times2},\] where each element of \( \mathbf{V}^0_{\text{occ},\text{max}} \) corresponds to \( v^0_{\text{occ},s} \), and each element of \( \mathbf{V}^1_{\text{occ},\text{max}} \) corresponds to \( v^1_{\text{occ},s} \).

The position, acceleration, and jerk constraints in both the longitudinal and lateral directions are enforced by \eqref{eq1:problem8} and \eqref{eq1:problem9}, respectively. Additionally, the consistency constraints in \eqref{eq1:problem10} and \eqref{eq1:problem11} ensure that the exploration and fallback trajectories share a common segment of length \( N_s \) in position, velocity, and orientation. This alignment is essential for maintaining a consistent driving policy in receding horizon planning, as the planned trajectory at the next step to be executed is shared by both the exploration and fallback trajectories.  

Note that the structure of the objective function \eqref{eq:obj_func_new} with the separable nature of the constraints in \eqref{eq1:problem} allows us to decompose the biconvex NLP problem through ADMM iterations.

\begin{remark}
{The proposed planner enhances {driving stability by generating smooth trajectories with consistency constraints. Specifically, smooth B\'ezier curves enforce bounded curvature and jerk, while the consistency constraints \eqref{eq1:problem10}-\eqref{eq1:problem11} prevent discontinuous jumps during trajectory switching.} In addition, the reachable set-based constraints guarantee that the fallback trajectory remains within safe dynamic limits. This prevents abrupt control changes during maneuvers and further enhances overall driving stability.}
\end{remark}

\subsubsection{Augmented Lagrangian} The augmented Lagrangian of the problem \eqref{eq1:problem} integrates both primal and dual variables, facilitating the enforcement of constraints and enhancing convergence within the ADMM framework. This formulation includes control points for orientation ($ \mathbf{C}_{\theta} $), longitudinal ($\mathbf{C}_x$), and lateral ($\mathbf{C}_y$) movements, along with slack variables ($\mathbf{S}_x,\mathbf{S}_y$) to handle inequality constraints, and consistency variables ($\mathbf{Z}_x,\mathbf{Z}_y ,\mathbf{Z}_\theta$) to ensure trajectory alignment between exploration and fallback trajectories. 
The augmented Lagrangian is given by 
\begin{align} \vspace{-0mm} 
&\mathcal{L}_{A}\Big(  { \mathbf{C}_{\theta} , \mathbf{C}_x, \mathbf{C}_y, }  {\mathbf{S}_x, \mathbf{S}_y},   {\bm{\omega}, \bm{\xi}},  \mathbf{Z}_x,\mathbf{Z}_y,\mathbf{Z}_{\theta} \nonumber\\
\hspace{-5mm}& \quad   \quad {\bm{\lambda}_{\theta}, \bm{\lambda}_{x}, \bm{\lambda}_{y},  \bm{\lambda}_{\text{obs},x}, \bm{\lambda}_{\text{obs},y}, \bm{\lambda}_{\text{c},x}, \bm{\lambda}_{\text{c},y}, \bm{\lambda}_{\text{c},\theta}} 
\Big)  \nonumber\\
    &= \mathcal{L}+  \mathcal{I}_{+}(\mathbf{S}_x) + \mathcal{I}_{+}(\mathbf{S}_y)  + \bm{\lambda}^T_{x}\mathbf{C}_x  + \bm{\lambda}^T_{y}\mathbf{C}_y \nonumber \\
 &\quad+ \bm{\lambda}^T_{\theta}  \left( \dot{\mathbf{W}}^T_B \mathbf{C}_{x}  - \mathbf{V} \cdot \cos{ (\mathbf{W}^T_B\mathbf{C}_{\theta})}  \right) \nonumber \\
    &\quad+ \bm{\lambda}^T_{\theta}  \left( \dot{\mathbf{W}}^T_B \mathbf{C}_{y}  - \mathbf{V} \cdot \sin{ (\mathbf{W}^T_B\mathbf{C}_{\theta})} \right) \nonumber\\
    &\quad+ \bm{\lambda}^T_{\text{obs},x}  \left(   \mathbf{A}_h \mathbf{C}_{x}  - \mathbf{O}_{x}  - \mathbf{L}_x \cdot  \bm{\xi}  \cdot  \cos ( \bm{\omega} )  \right) \nonumber\\
    &\quad+ \bm{\lambda}^T_{\text{obs},y}  \left(   \mathbf{A}_h \mathbf{C}_{y}  - \mathbf{O}_{y}  - \mathbf{L}_y \cdot  \bm{\xi}  \cdot  \sin ( \bm{\omega} )  \right) \nonumber\\
   &\quad+ \bm{\lambda}_{\text{c},x }  \left(  \mathbf{A}^T_{\text{c},x} \mathbf{C}_x    - \mathbf{Z}_{x}   \right)  + \bm{\lambda}_{\text{c},y }  \left(  \mathbf{A}^T_{\text{c},y} \mathbf{C}_y    - \mathbf{Z}_{y}   \right) \nonumber\\
   &\quad+ \bm{\lambda}_{\text{c},\theta }  \left(  \mathbf{A}^T_{\text{c},\theta} \mathbf{C}_\theta    - \mathbf{Z}_{\theta}   \right)  + \frac{\rho_x}{2} \left\| \mathbf{G}_x \mathbf{C}_x  -\mathbf{F}_x + \mathbf{S}_x \right\|_{2}^{2} \nonumber\\
    &\quad+ \frac{\rho_y}{2} \left\| \mathbf{G}_y \mathbf{C}_y  -\mathbf{F}_y + \mathbf{S}_y \right\|_{2}^{2} +  \frac{\rho_{\text{c},\theta }}{2}  \left\| \mathbf{A}^T_{\text{c},\theta}  \mathbf{C}_{\theta}     - \mathbf{Z}_{\theta} 
    \right\|_{2}^{2}  
    \nonumber\\ 
        &\quad+ \frac{\rho_{\text{c},x }}{2}  \left\| \mathbf{A}^T_{\text{c},x}  \mathbf{C}_x   - \mathbf{Z}_{x} 
    \right\|_{2}^{2}  + \frac{\rho_{\text{c}, y }}{2}  \left\| \mathbf{A}^T_{\text{c},y} \mathbf{C}_y    - \mathbf{Z}_{y} 
    \right\|_{2}^{2} 
\nonumber \\
    &\quad+ \frac{\rho_{\theta}}{2}  \left\| \dot{\mathbf{W}}^T_B \mathbf{C}_{x}  - \mathbf{V} \cdot \cos{ (\mathbf{W}^T_B\mathbf{C}_{\theta})}  
    \right\|_{2}^{2} \nonumber\\
    &\quad+ \frac{\rho_{\theta}}{2}  \left\| \dot{\mathbf{W}}^T_B \mathbf{C}_{y}  - \mathbf{V} \cdot \sin{ (\mathbf{W}^T_B\mathbf{C}_{\theta})}  
    \right\|_{2}^{2} \nonumber \end{align}\begin{align} 
    &\quad+ \frac{\rho_{\text{obs}}}{2}  \left\| 
    \mathbf{A}_h \mathbf{C}_{x}  - \mathbf{O}_{x}  - \mathbf{L}_x \cdot  \bm{\xi}  \cdot  \cos ( \bm{\omega} ) 
    \right\|_{2}^{2} 
       \nonumber\\
    &\quad+ \frac{\rho_{\text{obs}}}{2}  \left\|    \mathbf{A}_h \mathbf{C}_{y}  - \mathbf{O}_{y} - \mathbf{L}_y \cdot  \bm{\xi}  \cdot  \sin ( \bm{\omega} )  \right\|_{2}^{2},
    \label{lang_dual_problem}
\vspace{-0mm} \end{align}
where 
\( \rho_\theta,\rho_x,\rho_y, \rho_{\text{c},x}, \rho_{\text{c},y} , \rho_{\text{c},\theta} , \rho_{\text{obs}} \) are the penalty parameters and \(\bm{\lambda}_\theta, \bm{\lambda}_x, \bm{\lambda}_y,  \bm{\lambda}_{\text{c},x}, \bm{\lambda}_{\text{c},y}, \bm{\lambda}_{\text{c},\theta},  \bm{\lambda}_{\text{obs},x},  \bm{\lambda}_{\text{obs},y}\)  are dual variables associated with the respective constraints. $\mathbf{S}_x$ and $\mathbf{S}_y$ denote slack variables to transform the inequality constraints \(\eqref{eq1:problem7}\)-\(\eqref{eq1:problem9}\) into equality constraints
$\mathcal{I}_{+}(\cdot)$ denotes the indicator function of the nonnegative orthant, enforcing non-negativity of the slack variables, and is defined as
\begin{align}
    \mathcal{I}_{+}(\mathbf{Z}) =
    \begin{cases}
        0, & \text{if } \mathbf{Z} \ge 0 \text{ elementwise}, \\
        +\infty, & \text{otherwise},
    \end{cases}
\end{align} 
The matrix $ \mathbf{G}_y$ is a linear mapping matrix, given by 
  \begin{align}  
  \mathbf{G}_y = \mathbf{G} =   \big [\mathbf{W}^T_B\  -\mathbf{W}^T_B\  \ddot{\mathbf{W}}^T_B\  -\ddot{\mathbf{W}}^T_B\ \dddot{\mathbf{W}}^T_B\  -\dddot{\mathbf{W}}^T_B  
   \big]^T. \notag
\end{align} Since the matrix \( \mathbf{G}_x \) is designed to augment the coefficient matrix on the left-hand side of constraint \(\eqref{eq1:problem7}\), it expands the system in a structured manner, facilitating alternate optimization for the primal variable $\mathbf{C}_x$:
 \begin{align}   \vspace{-2mm}
  \mathbf{G}_x =  \big[
     &\mathbf{W}^T_B \quad -\mathbf{W}^T_B \quad \dot{\mathbf{W}}^T_B \quad -\dot{\mathbf{W}}^T_B \notag\\
    &\ddot{\mathbf{W}}^T_B \quad -\ddot{\mathbf{W}}^T_B \quad \dddot{\mathbf{W}}^T_B \quad -\dddot{\mathbf{W}}^T_B
    \big]^T.  \notag
  \end{align}   \vspace{-2mm}
\subsection{Consensus ADMM Iterations} \label{sec:ADMMiter}
\subsubsection{Update of Primal Variables  }
Following the ADMM iteration in~\cite{zheng2025barrier}, we begin by updating the control points for orientation, longitudinal, and lateral movements. The iterations proceed through the following steps:

\noindent\textbf{Step 1: Update of control point variables.}
The subproblem for the variable $\mathbf{C}_\theta$ is given by
\begin{subequations}\label{eq:sub_qp1}
\begin{align} \hspace{-2mm}
    \mathbf{C}^{\iota+1}_{\theta}  &:= \displaystyle\operatorname*{ \text{argmin}}_{\mathbf{C}_{\theta}}~~
    \mathcal{L}_A   \Big(  \mathbf{C}_\theta, \mathbf{C}^{\iota}_x, \mathbf{C}^{\iota}_y, \mathbf{S}^{\iota}_x,  \mathbf{S}^{\iota}_y,  \ \bm{\omega}^{\iota} ,  \bm{\xi}^{\iota} , \nonumber \\
   &\qquad\qquad\qquad~~~     \mathbf{Z}^{\iota}_x, \mathbf{Z}^{\iota}_y, \mathbf{Z}^{\iota}_{\theta} , \bm{\lambda}^{\iota}_{\theta},  \bm{\lambda}^{\iota}_{x}, \bm{\lambda}^{\iota}_{y} , \nonumber \\ 
    &\qquad\qquad\qquad~~~   \bm{\lambda}^{\iota}_{\text{obs},x}, \bm{\lambda}^{\iota}_{\text{obs},y} \bm{\lambda}^{\iota}_{\text{c},x}, \bm{\lambda}^{\iota}_{\text{c},y}, \bm{\lambda}^{\iota}_{\text{c},\theta}   \Big) \label{eq:sub_qp1_obj}\\
   &   \qquad \text{s.t.}\quad
    [\mathbf{A}_0 \quad \mathbf{A}_{f,\theta} ]^T \mathbf{C}_{\theta}  = [\bm{\theta}_0\quad \dot{\bm{\theta}}_0\quad \mathbf{0}\quad \mathbf{0}]^T,  \label{eq:sub_qp1_cons} 
 \end{align} 
\end{subequations}
where $\iota$ denotes the index of iterations. Since the subproblem is a convex quadratic programming (QP) problem with full-rank matrices $\mathbf{A}_0$ and $\mathbf{A}_{f,\theta}$, updating $\mathbf{C}_\theta$ reduces to solving a convex equality-constrained QP. The optimality condition for this subproblem is given by 
\begin{equation} \vspace{0mm}
   \mathbf{A}_{\theta} \mathbf{C}_{\theta} = \mathbf{b}_{\theta},
   \label{eq:theta_results}
 \vspace{-2mm} 
\end{equation}  
where 
\[
\mathbf{A}_{\theta}
= \begin{bmatrix}
\mathbf{Q}_{\theta}
+ \rho_{\theta}\mathbf{W}_B\mathbf{W}^T_B
+ \rho_{\text{c},\theta}\mathbf{A}_{\text{c},\theta}
\mathbf{A}^T_{\text{c},\theta}\\
\mathbf{A}_0 \\
\mathbf{A}_{f,\theta}
\end{bmatrix},
\quad
\mathbf{b}_{\theta}
=
\begin{bmatrix}
\mathbf{b}_{\theta,0}\\
\bm{\theta}_0\\
\dot{\bm{\theta}}_0\\
\mathbf{0}\\
\mathbf{0}
\end{bmatrix},
\]
with
\[
\begin{aligned}
\mathbf{b}_{\theta,0}
=&\;
\mathbf{W}_B
\left(
\rho_{\theta}\arctan\!\left(
{\dot{\mathbf{W}}^T_B \mathbf{C}^{\iota}_{y}}/
{\dot{\mathbf{W}}^T_B \mathbf{C}^{\iota}_{x}}
\right)
-\bm{\lambda}_{\theta}^{\iota}
\right)\\
&
- \mathbf{A}_{\text{c},\theta}\bm{\lambda}_{\text{c},\theta}^{\iota} +
\rho_{\text{c},\theta}\mathbf{A}_{\text{c},\theta}
\mathbf{Z}^{\iota}_{\theta}.
\end{aligned}
\]

Following the update of $\mathbf{C}_\theta$, we address the subproblem for updating $\mathbf{C}_x$:
\begin{subequations}\label{eq:sub_qp2}
\begin{align}  \vspace{-0mm}
    \mathbf{C}^{\iota+1}_{x}  &:= \displaystyle\operatorname*{ \text{argmin}}_{\mathbf{C}_{x}}~~
     \mathcal{L}_A   \Big(  \mathbf{C}^{\iota}_\theta, \mathbf{C}_x, \mathbf{C}^{\iota}_y, \mathbf{S}^{\iota}_x,  \mathbf{S}^{\iota}_y,  \ \bm{\omega}^{\iota} ,  \bm{\xi}^{\iota} , \nonumber \\
  &\qquad\qquad\qquad~~~     \mathbf{Z}^{\iota}_x, \mathbf{Z}^{\iota}_y, \mathbf{Z}^{\iota}_{\theta} , \bm{\lambda}^{\iota}_{\theta},  \bm{\lambda}^{\iota}_{x}, \bm{\lambda}^{\iota}_{y} , \nonumber \\ 
    &\qquad\qquad\qquad~~~   \bm{\lambda}^{\iota}_{\text{obs},x}, \bm{\lambda}^{\iota}_{\text{obs},y} \bm{\lambda}^{\iota}_{\text{c},x}, \bm{\lambda}^{\iota}_{\text{c},y}, \bm{\lambda}^{\iota}_{\text{c},\theta}   \Big) \label{eq:sub_qp2_obj}\\
   &   \qquad \text{s.t.}\quad
 \mathbf{A}^T_0  \mathbf{C}_{x}  = [\mathbf{P}_{x,0}\quad \dot{\mathbf{P}}_{x,0} ]^T,   \label{eq:sub_qp2_cons} 
\vspace{-0mm} \end{align}
\end{subequations}
where $\mathbf{P}_{x,0} \in \mathbb{R}^{2}$ and  $\dot{\mathbf{P}}_{x,0}\in\mathbb{R}^{ 2}$ represent the initial longitudinal position and velocity for the exploration and fallback trajectories, respectively. Given that \eqref{eq:sub_qp2_cons} is an affine equality constraint and that the weight matrices \( \mathbf{Q}_x \) and \( \mathbf{Q}_1 \) in \eqref{eq:obj_func_new_x} are positive semidefinite, the optimality condition for this convex problem is derived as
\begin{equation} \vspace{-0mm}
  \mathbf{A}_{x}  \mathbf{C}^{\iota+1}_{x} = \mathbf{b}_{x}, \label{eq:x_results}
  \end{equation} 
where  
\[ \bm{A}_{x} = \begin{bmatrix}   \bm{A}_{x,0} 
\\ \mathbf{A}_0  \end{bmatrix},  \mathbf{b}_{x} = \begin{bmatrix} 
  \mathbf{b}_{x,0}
  \\  \mathbf{P}_{x,0}
  \\\dot{\mathbf{P}}_{x,0}   \end{bmatrix}, \] 
and the matrices $\mathbf{A}_{x,0} $ and $\mathbf{b}_{x,0} $ are given by 
\begin{alignat}{2} \vspace{-0mm}
\mathbf{A}_{x,0} = & \mathbf{Q}_{x} + \dot{\mathbf{W}}_B  \mathbf{Q}_1 \dot{\mathbf{W}}_B^T + \rho_{\theta}\dot{\mathbf{W}}_B\dot{\mathbf{W}}^T_B 
+ \rho_{\text{obs}}\mathbf{A}^T_h  \mathbf{A}_h\nonumber \\
    & + \rho_{\text{c},x}\mathbf{A}_{\text{c},x}\mathbf{A}^T_{\text{c},x}
+ \rho_{x}  \mathbf{G}^T_x  \mathbf{G}_x ,\notag 
\vspace{-0mm} \end{alignat}    
\begin{alignat}{2} \vspace{-0mm}
 \mathbf{b}_{x,0} = &   \mathbf{V}_{x,d}^T \mathbf{Q}_1 \dot{\mathbf{W}}_B^T  -\bm{\lambda}_{x}^{\iota}
-\dot{\mathbf{W}}_B\bm{\lambda}_{\theta}^{\iota} 
- \mathbf{A}^T_h\bm{\lambda}_{\text{obs},x}^{\iota } \nonumber \\
    & +\rho_{\theta}\dot{\mathbf{W}}_B \mathbf{V} \cdot \cos{( \mathbf{W}^T_B\mathbf{C}^{\iota}_{\theta}} ) 
 \notag\\
& + \rho_{\text{obs}}\mathbf{A}^T_h ( \mathbf{O}_{x}  + \mathbf{L}_x \cdot  \bm{\xi}^{\iota} \cdot  \cos{(\bm{\omega}^{\iota})} ) \notag\\
 &  + \frac{\rho_x}{2} \mathbf{G}_x^T (\mathbf{F}_x-\mathbf{S}^{\iota}_x )+ \rho_{\text{c},x}\mathbf{A}_{\text{c},x}  \mathbf{Z}^{\iota}_{x}   .\notag 
\vspace{-15mm} \end{alignat} 

Similarly, the subproblems for updating the variable $\mathbf{C}_y$ is given by \begin{subequations}\label{eq:sub_qp3}
\begin{align} \hspace{-2mm}
    \mathbf{C}^{\iota+1}_{y}  &:= \displaystyle\operatorname*{ \text{argmin}}_{\mathbf{C}_{y}}~~
    \mathcal{L}_A   \Big(  \mathbf{C}^{\iota}_\theta, \mathbf{C}^{\iota}_x, \mathbf{C}_y, \mathbf{S}^{\iota}_x,  \mathbf{S}^{\iota}_y,  \ \bm{\omega}^{\iota} ,  \bm{\xi}^{\iota} , \nonumber \\
   &\qquad\qquad\qquad~~~     \mathbf{Z}^{\iota}_x, \mathbf{Z}^{\iota}_y, \mathbf{Z}^{\iota}_{\theta} , \bm{\lambda}^{\iota}_{\theta},  \bm{\lambda}^{\iota}_{x}, \bm{\lambda}^{\iota}_{y} , \nonumber \\ 
    &\qquad\qquad\qquad~~~   \bm{\lambda}^{\iota}_{\text{obs},x}, \bm{\lambda}^{\iota}_{\text{obs},y} \bm{\lambda}^{\iota}_{\text{c},x}, \bm{\lambda}^{\iota}_{\text{c},y}, \bm{\lambda}^{\iota}_{\text{c},\theta}   \Big) \label{eq:sub_qp3_obj}\\
   &   \qquad \text{s.t.}\quad
    [\mathbf{A}_0 \quad \mathbf{A}_{f,y} ]^T \mathbf{C}_{y}  = [\mathbf{P}_{y,0}\quad \dot{\mathbf{P}}_{y,0}\quad \mathbf{P}_{y,N} ]^T,  \label{eq:sub_qp3_cons} 
 \end{align} 
\end{subequations}
where the matrix $\mathbf{P}_{y,0} \in \mathbb{R}^{ 2}$ and  $\dot{\mathbf{P}}_{y,0}\in\mathbb{R}^{ 2}$ denote the initial lateral position and velocity for the exploration trajectory and the fallback trajectory, respectively. Additionally, $ \mathbf{P}_{y,N} \in\mathbb{R}^{ 2}$ ensures that the final lateral position of both trajectories aligns with the desired centerline. 
The optimality condition of \eqref{eq:sub_qp3} is expressed as 
\begin{equation} \vspace{-0mm}
  \mathbf{A}_{y}\mathbf{C}^{\iota+1}_{y} =    \mathbf{b}_{y}, 
 \label{eq:y_results} \end{equation} 
where 
$\bm{A}_{y} =  \begin{bmatrix}   \bm{A}_{y,0} 
\quad  \mathbf{A}_{0} \quad  \mathbf{A}_{f,y} \end{bmatrix}^T,  \mathbf{b}_{y} = \begin{bmatrix}  $ 
 $   \mathbf{b}_{y,0}
 \quad  \mathbf{P}_{y,0} \quad \dot{\mathbf{P}}_{y,0} \quad \mathbf{P}_{y,g}\end{bmatrix}^T, $ 
and the matrices $\mathbf{A}_{y,0} $ and $\mathbf{b}_{y,0} $ are given by
\begin{alignat}{2} \vspace{-0mm}
\mathbf{A}_{y,0} = & \mathbf{Q}_{y} + \mathbf{A}_{\text{y},\text{d}} \mathbf{Q}_2 \mathbf{A}^T_{\text{y},\text{d}} + \rho_{\theta}\dot{\mathbf{W}}_B\dot{\mathbf{W}}^T_B 
+ \rho_{\text{obs}}\mathbf{A}^T_h  \mathbf{A}_h\nonumber \\
    & + \rho_{\text{c},y}\mathbf{A}_{\text{c},y}\mathbf{A}^T_{\text{c},y}
+ \rho_{y}  \mathbf{G}^T_y  \mathbf{G}_y ,\notag 
\vspace{-0mm} \end{alignat}    
\begin{alignat}{2} \vspace{-0mm} 
 \mathbf{b}_{y,0} = & -\bm{\lambda}_{y}^{\iota}+ \mathbf{P}_{y,d}^T \mathbf{Q}_2 \mathbf{A}^T_{\text{y},\text{d}}
-\dot{\mathbf{W}}_B\bm{\lambda}_{\theta}^{\iota} 
- \mathbf{A}^T_h\bm{\lambda}_{\text{obs},y}^{\iota }  \notag\\
& +\rho_{\theta}\dot{\mathbf{W}}_B \mathbf{V} \cdot \sin{ (\mathbf{W}^T_B\mathbf{C}^{\iota}_{\theta})}  
 \notag\\
& + \rho_{\text{obs}}\mathbf{A}^T_h ( \mathbf{O}_{y}  + \mathbf{L}_y \cdot  \bm{\xi}^{\iota} \cdot  \sin{(\bm{\omega}^{\iota})} ) \notag\\
 &   + \frac{\rho_y}{2} \mathbf{G}_y^T (\mathbf{F}_y-\mathbf{S}^{\iota}_y )+ \rho_{\text{c},y}\mathbf{A}_{\text{c},y}  \mathbf{Z}^{\iota}_{y} .  \notag 
\vspace{-15mm} \end{alignat}    
\noindent\textbf{Step 2: Update of the variables  $\bm{\omega}$ and $\bm{\xi}$.}
Exploiting the conditions \eqref{eq:polar_safety}-\eqref{eq:omega}, the update for variables $\bm{\omega}$ and $\bm{\xi}$ can be determined by
\begin{equation} \vspace{-0mm}
    \bm{\omega}^{\iota+1} = \arctan\left(\frac{\mathbf{L}_x \cdot (\mathbf{A}_h \mathbf{C}^{\iota+1}_{y} - \mathbf{O}_{y})}{\mathbf{L}_y \cdot (\mathbf{A}_h \mathbf{C}^{\iota+1}_{x} - \mathbf{O}_{x})}\right),
    \label{eq:omega_admm_update}
 \vspace{-0mm} \end{equation}\begin{equation} 
\bm{\xi}^{\iota+1} = \max\left(\mathbf{1}, 1 + (1-\bm{\alpha}) \cdot (\bm{\xi}^{\iota} - 1)\right)
\label{eq:d_admm_update}  
\end{equation}   
where $\bm{\alpha} \in  \rr^{(N\times M)\times 2}$ denotes the barrier coefficient matrices, with each element within the interval \( (0, 1) \). In this study, we adopt a linearly increasing strategy for \(\bm{\alpha}\), ranging from \(0.4\) to \(1\) across planning steps. This strategy strikes a balance between expanding the solution space and reducing driving conservativeness, as outlined in \cite{zheng2025barrier}. These updates ensure that \(\bm{\omega}\) and \(\bm{\xi}\) satisfy the safety and feasibility conditions required for obstacle avoidance and trajectory planning.  

  
\noindent\textbf{Step 3: Update of consistency variables.}  
Following the approach in \cite{boyd2011distributed},  the global consistency variables  \(\mathbf{Z}_x\), \(\mathbf{Z}_y\), and \(\mathbf{Z}_\theta\)  are updated as follows: 
 \begin{subequations} \vspace{-0mm}\small
    \begin{align}  
       \vspace{-2mm} \mathbf{Z}^{\iota+1}_{x}[:,0] &= \mathbf{Z}^{\iota+1}_{x}[:,1] = \frac{1}{2} \mathbf{A}^T_{\text{c},x} \left( \mathbf{C}^{\iota+1}_x[:,0] + \mathbf{C}^{\iota+1}_x[:,1] \right), \label{eq:consensus_x_admm_updatesimp} \\
     \vspace{-2mm}    \mathbf{Z}^{\iota+1}_{y}[:,0] &= \mathbf{Z}^{\iota+1}_{y}[:,1] = \frac{1}{2} \mathbf{A}^T_{\text{c},y} \left( \mathbf{C}^{\iota+1}_y[:,0] + \mathbf{C}^{\iota+1}_y[:,1] \right), \label{eq:consensus_y_admm_updatesimp} \\
     \vspace{-2mm}    \mathbf{Z}^{\iota+1}_{\theta}[:,0] &= \mathbf{Z}^{\iota+1}_{\theta}[:,1] = \frac{1}{2} \mathbf{A}^T_{\text{c},\theta} \left( \mathbf{C}^{\iota+1}_{\theta}[:,0] + \mathbf{C}^{\iota+1}_{\theta}[:,1] \right). \label{eq:consensus_theta_admm_update_simp}
    \end{align}      
\end{subequations}   
Here, the mapping matrices \(\mathbf{A}_{\text{c},x} \in \mathbb{R}^{((n+1) \times 3) \times 2}\), \(\mathbf{A}_{\text{c},y} \in \mathbb{R}^{((n+1) \times 3) \times 2}\), and \(\mathbf{A}_{\text{c},\theta} \in \mathbb{R}^{(n+1) \times 2}\) transform the control points into consistency variables \(\mathbf{Z}_x \in \mathbb{R}^{3N_s \times 2}\), \(\mathbf{Z}_y \in \mathbb{R}^{3N_s \times 2}\), and \(\mathbf{Z}_\theta \in \mathbb{R}^{N_s \times 2}\), respectively.  
\begin{remark}
    Coordinating the exploration and fallback trajectories ensures they are aligned in position, velocity, acceleration, and orientation during the initial \( N_s \) steps. This alignment is crucial for maintaining consistent vehicle behavior, particularly in occluded environments where precise coordination enhances safety and travel efficiency. 
\end{remark}

\noindent\textbf{Step 4: Update of slack variables.} 
The update for slack variables is given by 
\begin{subequations} 
\vspace{-1mm}
    \begin{align} 
        \mathbf{S}^{\iota+1}_x &= \max \left( \mathbf{0}, \mathbf{F}_x - \mathbf{G}_x \mathbf{C}^{\iota+1}_x \right), \label{eq:s_x_relaxedadmm_update} \\
        \mathbf{S}^{\iota+1}_y &= \max \left( \mathbf{0}, \mathbf{F}_y - \mathbf{G}_y \mathbf{C}^{\iota+1}_y \right). \label{eq:s_y_relaxedadmm_update}  
    \end{align}  \vspace{-1mm}
\end{subequations} 

This update ensures that the slack variables remain non-negative, facilitating the search for a feasible solution to the optimization
problem \eqref{eq1:problem}.

\begin{algorithm}[t]
\caption{
{Occlusion-Aware Contingency Planner}}\label{alg:occlusion_planner}
\begin{algorithmic}[1]
\State \textbf{Input}:  
\Statex \hspace{1em} - $\mathbf{Q}_{1}, \mathbf{Q}_{2}, \mathbf{Q}_{\theta}, \mathbf{Q}_{x}, \mathbf{Q}_{y}$: Cost weights;  
\Statex \hspace{1em} - $\rho_{\theta}, \rho_x, \rho_y, \rho_{\text{c},x}, \rho_{\text{c},y}, \rho_{\text{c},\theta}, \rho_{\text{obs}}$: Penalty weights; 
\Statex \hspace{1em} - $N, N_s, N_d$: Planning horizon parameters; 
\Statex \hspace{1em} - $\epsilon^{\text{pri}} = 0.1$: Primal residual threshold;
\Statex \hspace{1em} - $\iota_{\text{max}} = 200$: Maximum iterations; 
\Statex \hspace{1em} - $r_l$: Perception range of the EV;
\Statex \hspace{1em} - $v_{pv,\text{max}} $: Maximum Velocity of PVs;
\Statex \hspace{1em} - $ v_{\text{occ,min}}, v_{\text{occ,max}}, c_{\text{th,min}}, c_{\text{th,max}}$: Risk parameters;
\State \textbf{Initialize}: Control points $\mathbf{C}_x, \mathbf{C}_y, \mathbf{C}_{\theta}$ for $\mathbf{x}^{\text{e}}, \mathbf{x}^{\text{s}}$; 
\State \textbf{While} task not done \textbf{do}:  
\State \hspace{1em} Measure the state of the EV $\mathbf{x}_0$ and SVs $\{\mathbf{o}^{(i)}_0\}_{i=1}^M$ \\ 
\hspace{1em} within
 perception range;
\Statex \hspace{1em} \textbf{Risk Assessment}:
\State \hspace{2em} Compute risk \(r(s,d)\) via (13); 
\State \hspace{2em} Update velocity boundaries \(v_{\text{occ,s}}^0, v_{\text{occ,s}}^1\) via (14);
\Statex \hspace{1em} \textbf{Trajectory Optimization}:
\State \hspace{2em} \textbf{For} \(\iota \gets 0\) \textbf{to} \(\iota_{\text{max}}\) \textbf{do}:
\State \hspace{3em} Solve \eqref{eq:theta_results}, \eqref{eq:x_results}, \eqref{eq:y_results} for $\mathbf{C}_{\theta}, \mathbf{C}_x, \mathbf{C}_y$;
\State \hspace{3em} Update \(\boldsymbol{\omega}, \boldsymbol{\xi}\) via \eqref{eq:omega_admm_update}-\eqref{eq:d_admm_update};
\State \hspace{3em} Update \(\mathbf{Z}_x, \mathbf{Z}_y, \mathbf{Z}_{\theta}\) via \eqref{eq:consensus_x_admm_updatesimp}-\eqref{eq:consensus_theta_admm_update_simp};
\State \hspace{3em} Update slack variables \(\mathbf{S}_x, \mathbf{S}_y\) via \eqref{eq:s_x_relaxedadmm_update}-\eqref{eq:s_y_relaxedadmm_update};
\State \hspace{3em} Update dual variables $\bm{\lambda}_\theta$, $\bm{\lambda}_x$, $\bm{\lambda}_y$,  $\bm{\lambda}_{\text{c},x}$, $\bm{\lambda}_{\text{c},y}$, $\bm{\lambda}_{\text{c},\theta}$,  \\
 \hspace{3em}  $\bm{\lambda}_{\text{obs},x}$,  $\bm{\lambda}_{\text{obs},y}$  in parallel via \eqref{eq:lambda_theta_admm_update}-\eqref{eq:lambda_consensustheta_admm_update};
\State \hspace{3em} \textbf{Break if} primal residual \(< \epsilon^{\text{pri}}\);
\State \hspace{2em} \textbf{End For};
\State \hspace{1em} Apply the first \(N_s\) steps of the common trajectory  \\
\hspace{1em} segment to the EV via \eqref{eq:ev_model};
\State \textbf{End While};
\end{algorithmic} 
\end{algorithm} 

\subsubsection{Update of Dual Variables}   
\begin{subequations}
    \begin{align} 
         \bm{\lambda}^{\iota+1}_{\theta}  &= \bm{\lambda}^{\iota}_{\theta} + \rho_{\theta} \left(  \mathbf{W}^T_B\mathbf{C}^{\iota+1}_{\theta} -  
                 \arctan \left(\frac{ \dot{\mathbf{W}}^T_B\mathbf{C}^{\iota+1}_{y} }{  \dot{\mathbf{W}}^T_B\mathbf{C}^{\iota+1}_{x} }\right) \right),\label{eq:lambda_theta_admm_update} 
        \\
        \bm{\lambda}^{\iota+1}_x &=  \bm{\lambda}^{\iota}_x + \rho_{x} \left( \mathbf{G}_x \mathbf{C}^{\iota+1}_x  -\mathbf{F}_x + \mathbf{S}^{\iota+1}_x \right),\label{eq:lambda_x_admm_update}\\
        \bm{\lambda}^{\iota+1}_y &=  \bm{\lambda}^{\iota}_y + \rho_{y} \left( \mathbf{G}_y \mathbf{C}^{\iota+1}_y  -\mathbf{F}_y + \mathbf{S}^{\iota+1}_y \right),\label{eq:lambda_y_admm_update}  \end{align}   \begin{align}
        \bm{\lambda}^{\iota+1}_{\text{obs},x} &= \bm{\lambda}^{\iota}_{\text{obs},x} + \rho_{\text{obs}} \Bigl( \mathbf{A}_h\mathbf{C}^{\iota+1}_{x} - \mathbf{O}_{x}  \notag \\
        &\quad \quad \quad \quad  \quad  \quad   - \mathbf{L}_x \cdot  \bm{\xi}^{\iota+1} \cdot  \cos{(\bm{\omega}^{\iota+1})} \Bigr), \label{eq:lambda_obsx_admm_update} \\
        \bm{\lambda}^{\iota+1}_{\text{obs},y} &= \bm{\lambda}^{\iota}_{\text{obs},y} + \rho_{\text{obs}} \Bigl( \mathbf{A}_h\mathbf{C}^{\iota+1}_{y} - \mathbf{O}_{y} \notag \\
          &\quad \quad \quad \quad  \quad   \quad - \mathbf{L}_y \cdot  \bm{\xi}^{\iota+1} \cdot  \sin{(\bm{\omega}^{\iota+1})} \Bigr), \label{eq:lambda_obsy_admm_update} \\
        \bm{\lambda}^{\iota+1}_{\text{c},x} &=  \bm{\lambda}^{\iota}_{\text{c},x}  + \rho_{\text{c},x} \mathbf{G}_x^T \left( \mathbf{A}^T_{\text{c},x}  \mathbf{C}^{\iota+1}_x  -\mathbf{Z}^{\iota+1}_{x} \right),\label{eq:lambda_consensusx_admm_update}  \\
        \bm{\lambda}^{\iota+1}_{\text{c},y} &=  \bm{\lambda}^{\iota}_{\text{c},y}  + \rho_{\text{c},y} \mathbf{G}_y^T  \left( \mathbf{A}^T_{\text{c},y}  \mathbf{C}^{\iota+1}_y  -\mathbf{Z}^{\iota+1}_{y} \right),\label{eq:lambda_consensusy_admm_update} \\
        \bm{\lambda}^{\iota+1}_{\text{c},\theta} &=  \bm{\lambda}^{\iota}_{\text{c},\theta}  + \rho_{\text{c},\theta}  \left( \mathbf{A}^T_{\text{c},\theta}  \mathbf{C}^{\iota+1}_{\theta}  -\mathbf{Z}^{\iota+1}_{\theta} \right).\label{eq:lambda_consensustheta_admm_update}
    \end{align}   
\end{subequations}   



Based on our experimental results, the ADMM iterations terminate when the relative primal residual error $\epsilon^{\text{pri}}$ falls below a threshold of  0.1. This threshold ensures a balance between computational efficiency and solution accuracy, which is critical for real-time autonomous driving applications. If the condition is not met, the iteration index \(\iota\) is incremented, and the algorithm proceeds to Step 1. This process continues until the residual error drops below the threshold or the maximum number of iterations, \(\iota_{\text{max}} = 200\), is reached. {Detailed procedure of our occlusion-aware contingency planner is described in \textbf{Algorithm} \ref{alg:occlusion_planner}.}
 
 \subsubsection{Numerical Optimization} To enhance numerical stability, we employ the Householder QR decomposition~\cite{golub2013matrix} when solving the linear systems arising from the optimality conditions in equations \eqref{eq:theta_results}, \eqref{eq:x_results}, and \eqref{eq:y_results}. {Numerical stability ensures that small errors in computation do not propagate and significantly distort the final solution.} For example, consider the optimality condition in \eqref{eq:y_results}. Applying the Householder transformation, the matrix \(\mathbf{A}_y\) is decomposed into an orthogonal matrix \(\mathbf{Q}\) and an upper triangular matrix \(\mathbf{R}\):
\[
\mathbf{A}_y = \mathbf{Q} \mathbf{R}.
\]

This decomposition allows the equation to be rewritten as
\[
\mathbf{Q} \mathbf{R} \mathbf{C}^{\iota+1}_y = \mathbf{b}_y.
\]

By multiplying both sides by \(\mathbf{Q}^\top\), we leverage the orthogonal property of \(\mathbf{Q}\) to simplify the equation to 
\[
\mathbf{R} \mathbf{C}^{\iota+1}_y = \mathbf{Q}^\top \mathbf{b}_y.
\]

The upper triangular matrix \(\mathbf{R}\) facilitates the efficient computation of \(\mathbf{C}^{\iota+1}_y\) through back substitution. This method enhances the numerical stability of our computations and improves the accuracy of the solutions, particularly when dealing with ill-conditioned matrices~\cite{trefethen2022numerical}.

{
\begin{remark}
The dominant per-iteration computational costs arise from constraint-handling updates. Specifically:
\begin{itemize}
    \item The primal control point updates \eqref{eq:theta_results}, \eqref{eq:x_results}, \eqref{eq:y_results} exhibit $\mathcal{O}((n+1)^3)$ complexity. For fixed curve order $n=10$, this reduces to $\mathcal{O}(1)$ per iteration.
    \item Obstacle-related operations (Step 2) scale as $\mathcal{O}(N \cdot M)$, where $M$ is the number of obstacles.
    \item {Averaging operations over the consensus segment (Step 3) scale as $\mathcal{O}(N_s \cdot m)$, where $N_s$ is the consensus horizon and $m$ is the state dimension. For fixed parameters $N_s$ and $m$, this reduces to $\mathcal{O}(1)$ per iteration.}
    \item {The slack variable updates~\eqref{eq:s_x_relaxedadmm_update}~\eqref{eq:s_y_relaxedadmm_update} exhibit $\mathcal{O}(N \cdot c)$ complexity with element-wise $\max$ operation, where $c$ is the number of inequality constraints per step. }
    \item {The obstacle-related dual variable updates~\eqref{eq:lambda_obsx_admm_update}~\eqref{eq:lambda_obsy_admm_update} scale as $\mathcal{O}(N \cdot M)$. Dual variable updates~\eqref{eq:lambda_theta_admm_update}-\eqref{eq:lambda_y_admm_update} and~\eqref{eq:lambda_consensusx_admm_update}-\eqref{eq:lambda_consensustheta_admm_update} scale as $\mathcal{O}(N)$ and $\mathcal{O}(N_s)$, respectively.}
\end{itemize}
Given fixed parameters $n$ and $N_s$, and fixed horizon $N$ per planning cycle, the computational complexity per ADMM iteration scales linearly with the number of obstacles as $\mathcal{O}(M)$. The dual variable updates \eqref{eq:lambda_theta_admm_update}-\eqref{eq:lambda_consensustheta_admm_update} are executed in parallel. This architecture enables real-time iterations in dense traffic scenarios.
\end{remark}}

{\begin{remark}
The proposed planning approach maintains robustness in the presence of disturbances through three integrated mechanisms: (1) probabilistic occlusion modeling via spatial risk quantification handles sensing uncertainties, (2) slack-augmented consensus ADMM optimization temporarily accommodates perturbations while preserving hard constraints, and (3) dual-trajectory architecture ensures driving stability during transitions through smooth B\'ezier parameterization and consistency constraints. These mechanisms maintain stable performance under sudden environmental changes.
\end{remark}}
\vspace{-2mm}
\section{{Results}}\label{sec:Results}
In this section, we evaluate our contingency planning approach through detailed simulations and real-world experiments. We simulate an occluded intersection environment under dense traffic conditions. Moreover, real-world experiments demonstrate the practical applicability of our method using an Ackermann mobile robot platform.
\vspace{-2mm}
\subsection{Simulation}\label{subsec:Sim_results}
\subsubsection{Simulation Setup}
All simulations are conducted using C++ on the Ubuntu 20.04 LTS system, equipped with an AMD Ryzen 7 5800H CPU featuring eight cores and sixteen threads.  Visualization is performed using RViz in ROS Noetic at a communication rate of 100 Hz. The planning loop operates at 10 Hz with a discrete time step of $0.1\,\text{s}$. The planning step is set to \( N = 40 \)  with a time step of \( \Delta t = 0.1\,\text{s}\) for both exploration and fallback trajectories over the planning horizon. The consistency step is configured with \( N_s = 5 \) steps and a free step of \( N_d = 5 \) steps. 

The weighting matrices in \eqref{eq:obj_func_new} are set as \( \mathbf{Q}_{\theta} = 150 \), \( \mathbf{Q}_x = 100 \), \( \mathbf{Q}_y = 100 \), \( \mathbf{Q}_1 = 50 \), and \( \mathbf{Q}_2 = 100 \). All \( l_2 \) penalty parameters in \eqref{lang_dual_problem} are set to 5. The minimum risk threshold \( c_{\text{th,min}} \) is set to 0, while the maximum risk thresholds $c_{\text{th,max}}$ for the exploration and fallback trajectories are set to 60 and 40, respectively. The axes of the safe ellipse are configured as \( l_x = 0.6\,\text{m} \) and \( l_y = 0.6\,\text{m}\). Additional vehicle parameters are detailed in Table~\ref{table:Parameters}.

   \begin{table}[tp]
    \centering
    \scriptsize
    \setlength{\abovecaptionskip}{-0mm}
    \setlength{\belowcaptionskip}{-0mm}
    \caption{Vehicle Parameters} \vspace{0mm}
    \label{table:Parameters}
    \begin{tabular}{c c c}
    \hline \hline
    Description         &  Value   \\ \hline 
    Front axle distance to center of mass &   $l_f = 1.06\,\text{m}$\\
    Rear axle distance to center of mass &  $l_r =  1.85\,\text{m}$\\
    Maximum anticipated number of obstacles & $M=4$\\
    Maximum velocity of PVs & $v_{pv,\text{max}} = 10\,\text{m/s}$ \\
    Longitudinal position range  &  $p_{x}  \in[-50\,\text{m}, 50\,\text{m}] $\\
    Velocity range of SVs &  $  [0\,\text{m/s}, 10\,\text{m/s}] $\\     
    Acceleration range of SVs &  $  [-4\,\text{m/s}^2, 4\,\text{m/s}^2] $\\ 
    Velocity range of the EV &  $  [0\,\text{m/s}, 10\,\text{m/s}] $\\ 
    Perception range of the EV & $r_l =30\,\text{m} $\\ 
    Longitudinal acceleration range  &  $a_{x} \in [-6\,\text{m/s}^2, 4\,\text{m/s}^2] $\\
    Lateral acceleration range  &  $a_{y} \in [-3\,\text{m/s}^2, 3\,\text{m/s}^2] $\\
    Longitudinal jerk range  &  $j_{x} \in [-6\,\text{m/s}^3, 6\,\text{m/s}^3] $\\
    Lateral jerk range  &  $j_{y} \in [-6\,\text{m/s}^3, 6\,\text{m/s}^3] $\\
  \hline\hline
    \end{tabular}\vspace{-4mm}
    \end{table}  

\begin{figure}[tp]
    \centering
        \vspace{2mm}\includegraphics[width=.425\textwidth] {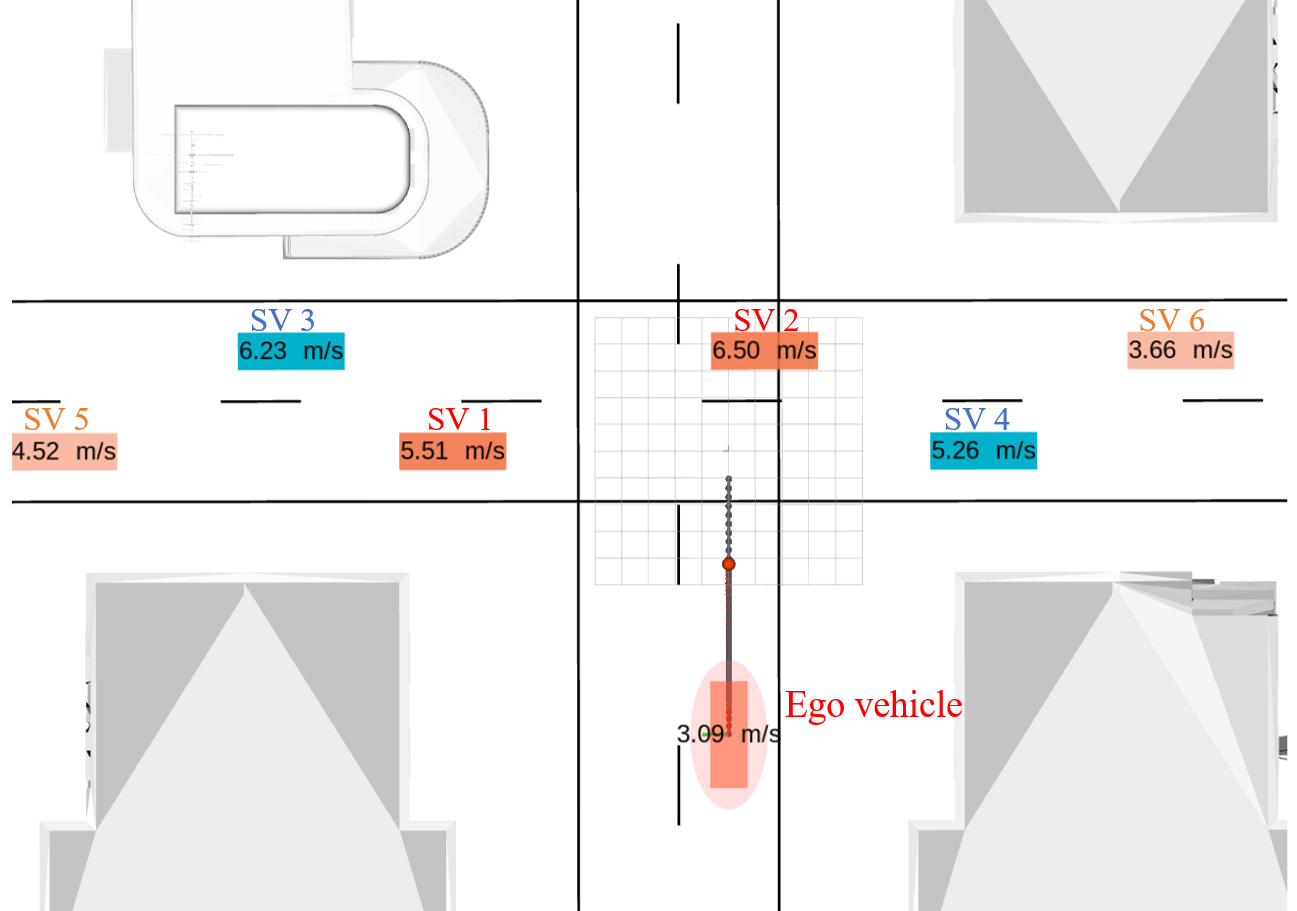}
    \caption{A top-down view of the EV navigating through a dynamic and occluded intersection under dense traffic conditions. The blue vehicles (SV3, SV4) represent non-interacting vehicles whose future trajectories do not intersect with the EV. The lighter orange vehicles (SV5, SV6) are occluded, while the solid orange vehicles (SV1, SV2) are fully visible. The EV plans two potential trajectories to manage the occlusion risks: the black exploration trajectory prioritizes travel efficiency, while the conservative fallback trajectory accounts for higher occlusion risks. }	
    \label{fig:sim_top-down}
    \vspace{-2mm}
\end{figure}

\begin{figure}[tp]
    \centering
       \hspace{-3.0mm}  \subfigure[ Time instant $t = 5.5\,\text{s}$]{
            \label{fig:shapshot_1} \hspace{0mm}
        \includegraphics[width=.234\textwidth]{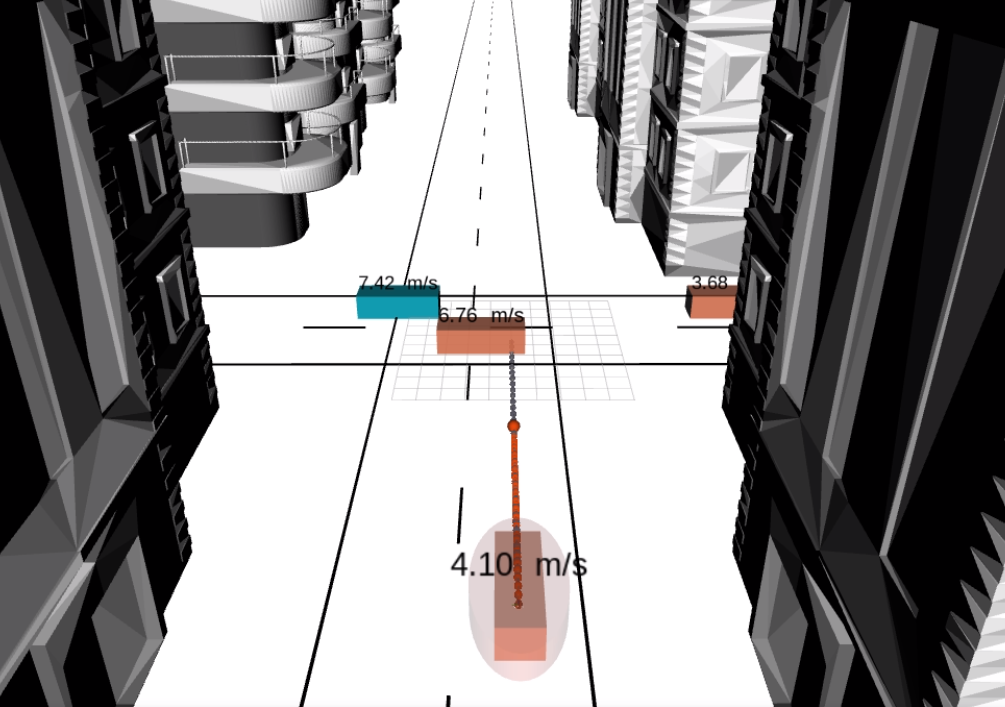}}
          \subfigure[ Time instant $t = 7\,\text{s}$]{
            \label{fig:shapshot_2}
        \includegraphics[width=.231\textwidth]{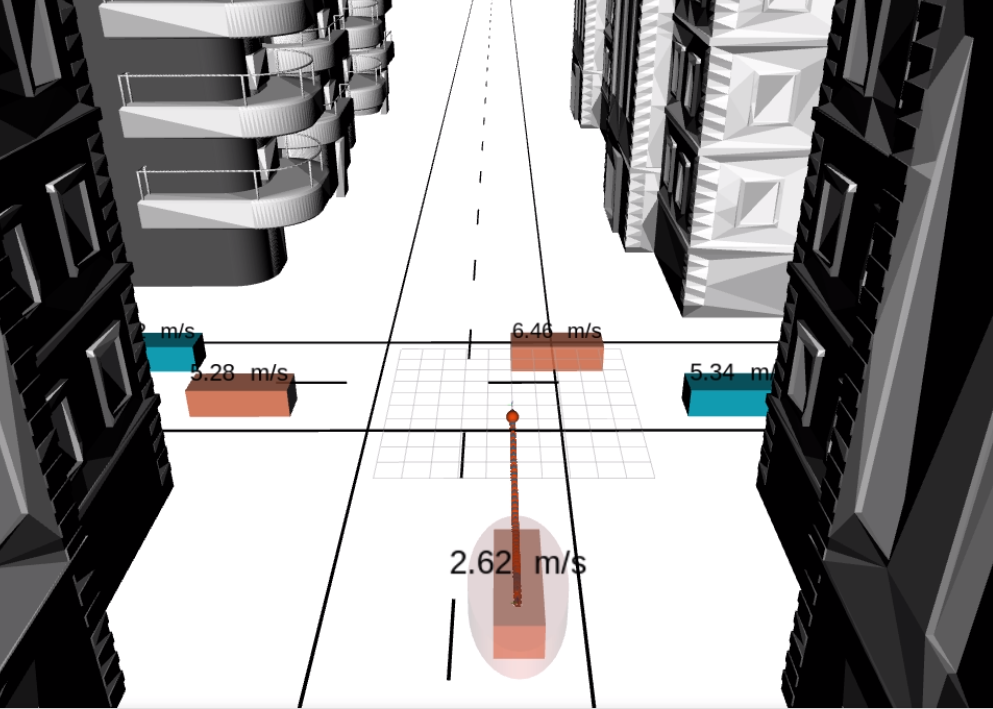}}\hspace{0mm}
          \subfigure[ Time instant $t = 9.9\,\text{s}$]{
            \label{fig:shapshot_3}
        \includegraphics[width=.234\textwidth]{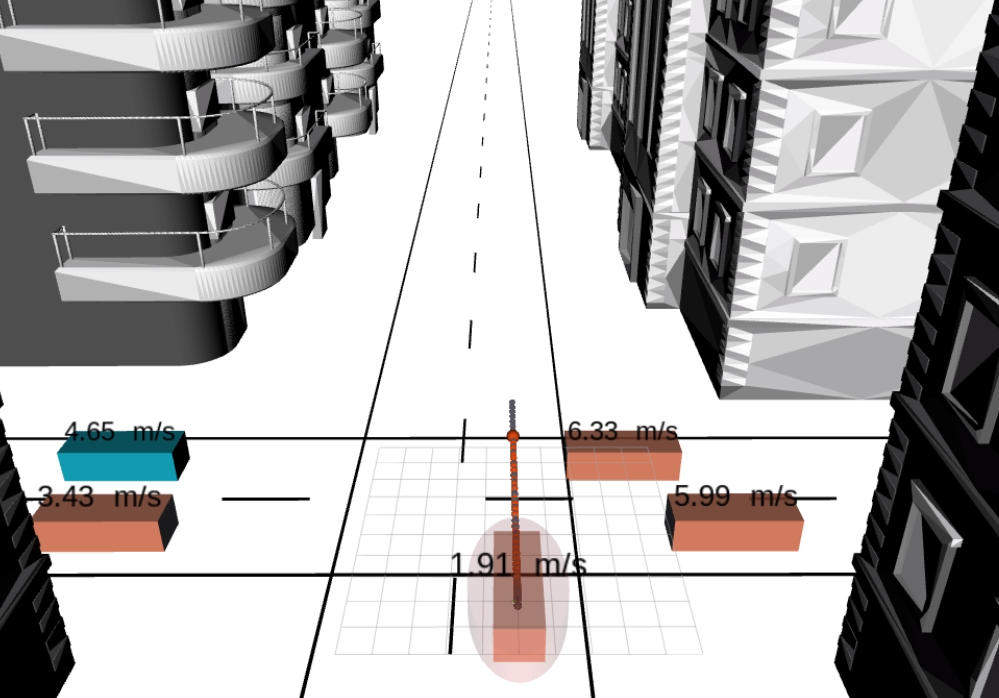}}\hspace{0mm}
          \subfigure[ Time instant $t = 12.5\,\text{s}$]{
            \label{fig:shapshot_4}
        \includegraphics[width=.234\textwidth]{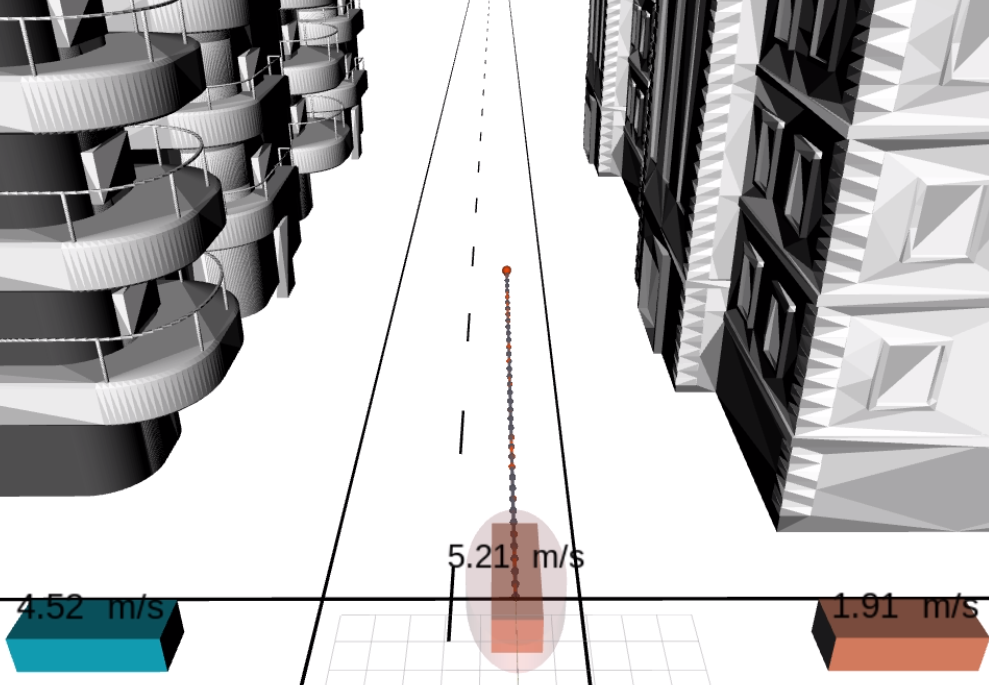}}\hspace{0mm}
    \vspace{-4mm}
    \caption{Third-person view snapshots of the red EV navigating through a dense and occluded intersection at different time instants. The proposed occlusion-aware contingency planner enables the EV to first decelerate to avoid potential hazards and then accelerate to safely pass through the occluded intersection under dense traffic conditions. }	
    \label{fig:sim_snapshots}
    \vspace{-4mm}
\end{figure}

In the simulated occluded intersection environment, as illustrated in Figs.~\ref{fig:sim_top-down} and \ref{fig:sim_snapshots}, ten SVs drive within two lanes traveling in opposite directions, adhering to right-hand traffic regulations over a longitudinal range from $-70\,\text{m}$ to $70\,\text{m}$. The longitudinal positions of the centerlines for the two lanes are $0\,\text{m}$ and $3.75\,\text{m}$, respectively. All SVs drive forward, following an intelligent driving model, with desired velocities ranging from $4\,\text{m/s}$ to $9.5\,\text{m/s}$. The safe time headway and distance are set to $1\,\text{s}$  and $3\,\text{m}$, respectively, to create a dense traffic condition. The target longitudinal velocity of the EV is \( v_{x,d} = 7\,\text{m/s} \) for both exploration and fallback trajectories. Starting from an initial position of \([-50\,\text{m}, 0\,\text{m}]^T\) and an initial velocity of  \([5\,\text{m/s}, 0\,\text{m/s}]^T\),  the EV is tasked with navigating through the occluded intersection while maintaining its original desired driving lane. 

\begin{table*}[t]

            \centering
            \scriptsize
            \setlength{\abovecaptionskip}{-0mm}
            \setlength{\belowcaptionskip}{-0mm}
            \caption{{Quantitative Results Comparison Among Different Algorithms} }
            \label{tab:comparasion_results}    
           \begin{tabular}[c]{|c | c | c| *{3}{c} | *{3}{c} |}
                \hline
                \multirow{2}{*}{\textbf{Algorithm}} & 
                 \multirow{2}{*}{{\textbf{Collision}}} &
                \multirow{2}{*}{\textbf{ {\textbf{Task Duration}}} (\text{s})} &
                \multicolumn{3}{c|}{{\textbf{Longitudinal Velocity}} (\text{m/s})} &   
                \multicolumn{3}{c|}{{\textbf{Solving Time}} ($\text{ms}$)} \\
               & &   & \efficiencycruise   & \efficiencycomputation    \\
                \hline
                        %
                \multirow{1}{*}{ Occlusion-Ignorant}
                 & \text{Yes}   & -- &--& -- & --  &-- & --& --   \\
                \hline           
                \multirow{1}{*}{{ST-RHC}} 
                & \text{No}   & 17.90  &  3.11 & 6.37 & 0.00    & 43.14
              & 227.06 &  5.21  \\
                \hline   
              \multirow{1}{*}{{Control-Tree} }
                & {\text{No}}   & {16.40}&  {3.53}& {7.00} & { 2.51$\times 10^{-3}$}    & {260.86}
              & {270.75} &  {240.25} \\
                \hline   
                \multirow{1}{*}{\textbf{Occlusion-Aware Contingency Planner}}
                 & \textbf{No}   &  \textbf{12.50}   & \textbf{4.68} &  \textbf{7.41} &\textbf{1.64} &  \textbf{23.83} &  \textbf{44.75} &\textbf{0.56}   \\ 
                \hline
            \end{tabular}    \vspace{-6mm}
        \end{table*} 
    
\textbf{Baselines.} { We compare our method against three baselines:  
\begin{itemize}  
    \item An occlusion-ignorant ablation of our algorithm;  
    \item Control-Tree approach~\cite{phiquepal2021control}, a branch MPC framework for autonomous driving in partially observable environments, implemented using its {open-source code\footnote{{\url{https://github.com/ControlTrees/icra2021}}}};  
    \item ST-RHC~\cite{zheng2024}, a nonlinear MPC (NMPC) approach using the ACADO toolkit~\cite{Houska2011a} with multiple shooting and sequential quadratic programming for efficient optimization.
\end{itemize}   }
{To ensure a fair comparison, parameter tuning is conducted for all baselines to achieve their best possible performance under consistent test conditions.} 

\textbf{Evaluation Metrics.} We assess performance using the following metrics: 
\begin{itemize} 
\item \textbf{Safety}: Whether a collision occurs during navigation. 
\item \textbf{Task Duration}: The time taken to traverse the occluded intersection from the initial position. 
\item \textbf{Computational Efficiency}: Optimization time to generate feasible trajectories. 

\end{itemize}

\begin{figure}[tp] 
\centering \includegraphics[width=.48\textwidth]{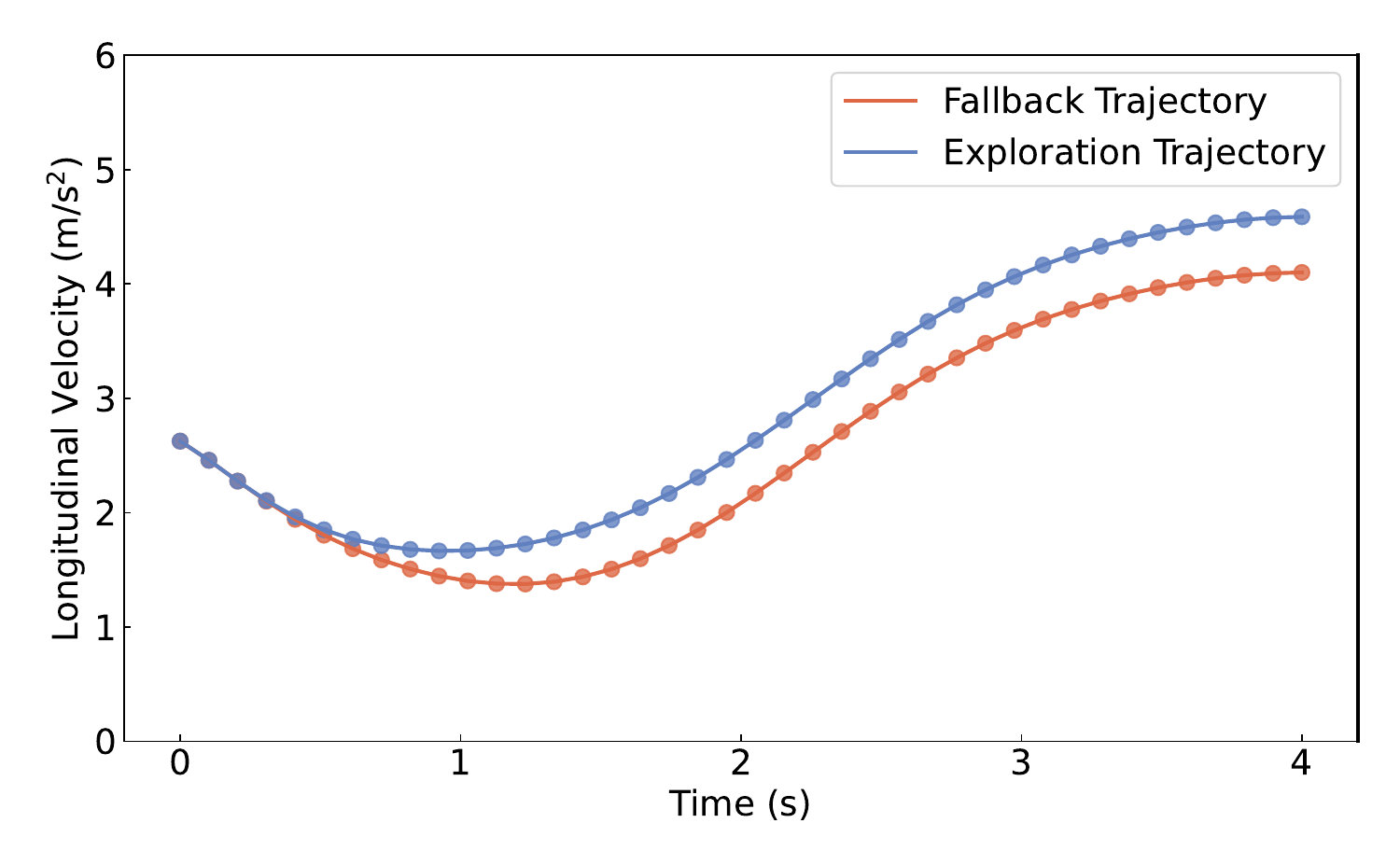} \vspace{-2mm} \caption{The generated velocity profiles of two trajectories at time instant $7\,\text{s}$. Two trajectories show the shared initial segment and the separate process for safe navigation.} \vspace{-4mm} \label{fig:Contingency_vel} \end{figure}

\vspace{-4mm}
\subsection{{Simulation Results}} 
Figure \ref{fig:sim_top-down} depicts a top-down view snapshot of the EV based on the proposed contingency planning approach. The EV approaches the intersection, with vehicles SV1 and SV2 fully visible, while SV5 and SV6 are occluded by buildings and SV4, respectively. The detailed evolution of the process is shown in Fig.~\ref{fig:sim_snapshots}. At time instant \( t = 5.5 \, \text{s} \), the EV is approaching the intersection. To address potential risks in this occluded intersection, the EV simultaneously plans two trajectories: the black exploration trajectory, which is more aggressive and designed to enhance travel efficiency while maintaining safety, and the red fallback trajectory, which prioritizes safety in response to increased occlusion. As a result, the EV chooses the fallback red trajectory and decelerates to allow the agent ahead to pass safely. By the time instant \( t = 12.5 \, \text{s} \), the EV has nearly completed its traversal of the intersection, with SVs becoming more visible and no vehicles ahead. At this stage, both the exploration trajectory and the fallback trajectory share the same risk value and converge to the same trajectory, as shown in Fig.~\ref{fig:shapshot_4}. The EV adjusts its trajectory based on updated environmental information and accelerates to a speed of $ 5.23  \, \text{m/s}$ to safely navigate the intersection. Notably, during this process, both trajectories share a common segment during the first five steps, as illustrated in Fig.~\ref{fig:Contingency_vel}. The velocity profiles of the exploration and fallback trajectories share a common initial segment at \( 7 \, \text{s} \), as observed in the receding horizon planning process.  This shared segment ensures a smooth and consistent driving experience, regardless of the trajectory the EV selects in subsequent steps, striking a balance between safety and travel efficiency. 

\begin{figure}[tb]
\begin{center}
\includegraphics[width=.45\textwidth]{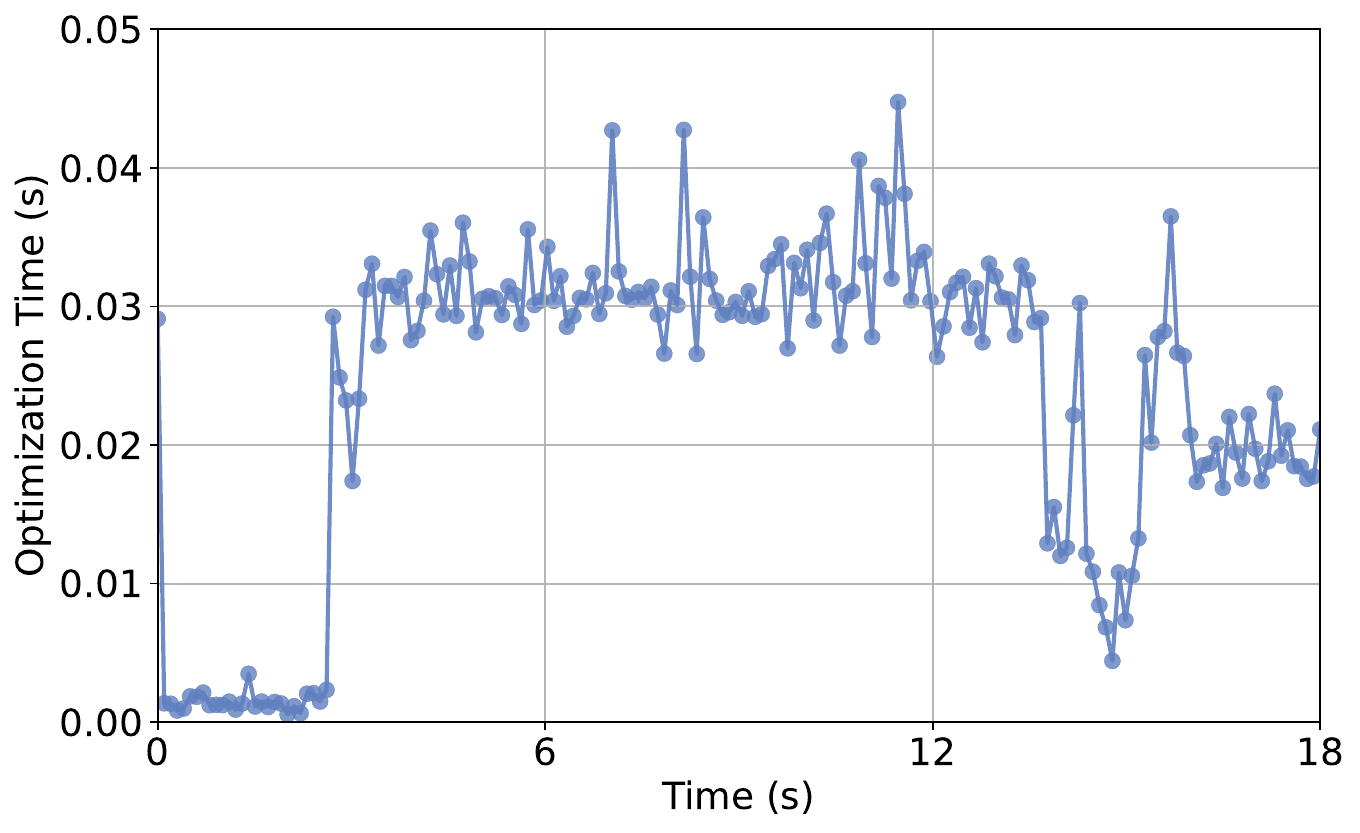}    \vspace{-3mm}
\caption{Evolution of optimization of the proposed occlusion-aware contingency planning approach with the prediction step $N=40$.} \vspace{-4mm}
\label{fig:Sim_time}
\end{center}
\end{figure}     

Table~\ref{tab:comparasion_results} presents the performance comparison among different approaches over a simulation time of $18\,\text{s}$. Notably, the occlusion-ignorant approach results in a collision with surrounding PVs at the intersection, underscoring the need to reason about unobserved space. Although the ST-RHC and Control-Tree approaches traverse the intersection safely with carefully tuned parameters, they take a longer time to pass through. Our planner shortens the intersection-traversal time to $12.50\,\text{s}$, achieving a 30.17\% and 23.78\% reduction compared with ST-RHC and Control-Tree, respectively. Likewise, the average cruise velocity improves by  50.48\% and  32.58\%  compared to ST-RHC ($4.68\,\text{m/s} $ versus  $3.11\,\text{m/s}$)  and Control-Tree ($4.68\,\text{m/s} $ versus  $3.53\,\text{m/s}$), respectively.

Notably, the minimum longitudinal velocity for ST-RHC and Control-Tree {decreases to $0\,\text{m/s}$ and $2.51\times 10^{-3}\,\text{m/s}$, respectively, whereas our contingency planner maintains at least $1.64\,\text{m/s}$. This indicates that the baseline methods adopt overly conservative braking to facilitate safe navigation in dynamic, occluded environments.}
 
Regarding computational efficiency, {the proposed occlusion-aware contingency planner requires only $23.83\,\text{ms}$ on average per optimization. This is 44.75\% faster than ST-RHC and 90.86\% faster than Control-Tree,  demonstrating real-time replanning capability.}  For a more intuitive view of this process, the evolution of the optimization time is shown in Fig.~\ref{fig:Sim_time}. Before approaching the intersection, the optimization time is fast, under $1\,\text{ms}$, but increases as the EV interacts with the SVs. After passing the intersection, the optimization time decreases. Detailed discussions on computational efficiency are provided in Section~\ref{sec:Discussion}.   

\subsection{Real-World Experiments}

\begin{figure}[tbp]
\centering
\hspace{-0mm} \includegraphics[width=.225\textwidth]{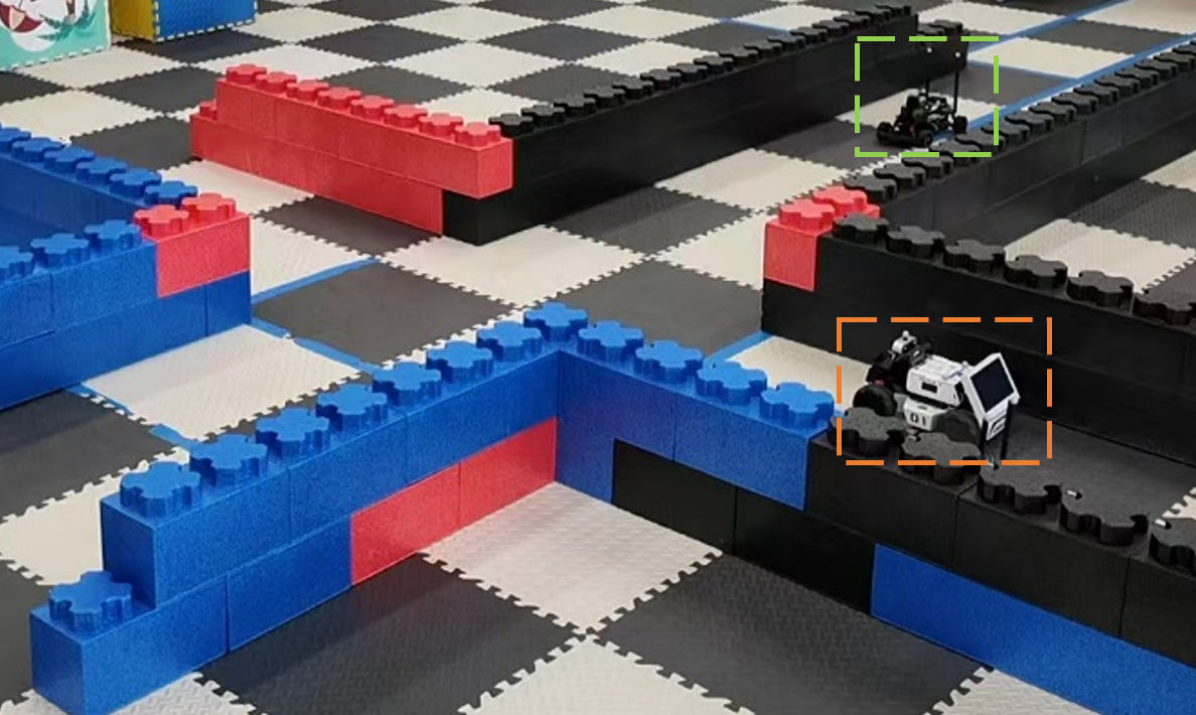}\hspace{2mm}\includegraphics[width=.225\textwidth]{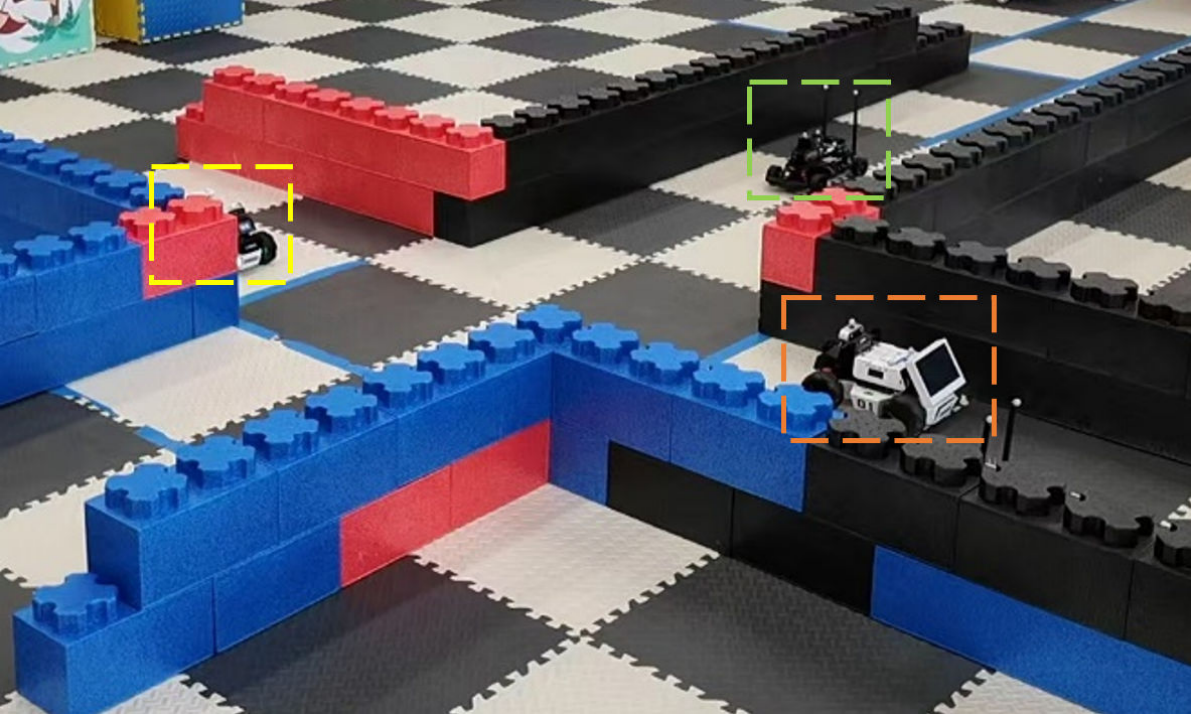}\\
\footnotesize{(a) The EV maintains a stable speed of approximately $ 0.5\,\text{m/s}$.} \\[5pt]
\hspace{-0mm}  \includegraphics[width=.225\textwidth]{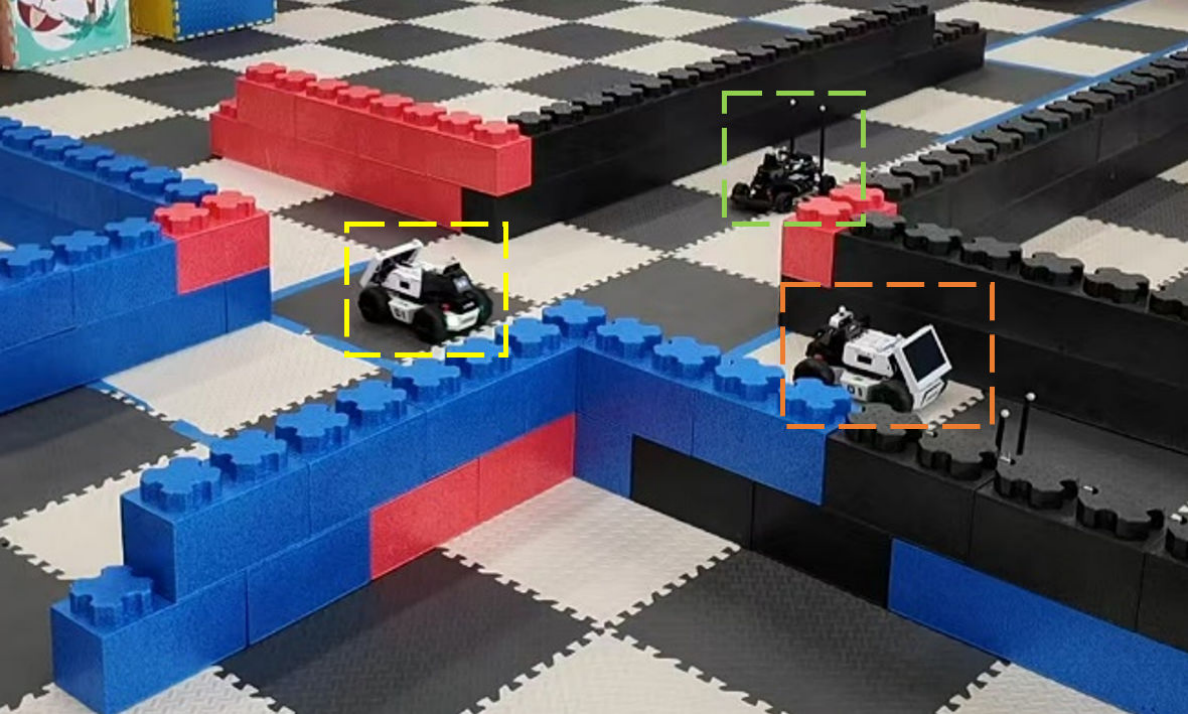}\hspace{2mm}\includegraphics[width=.225\textwidth]{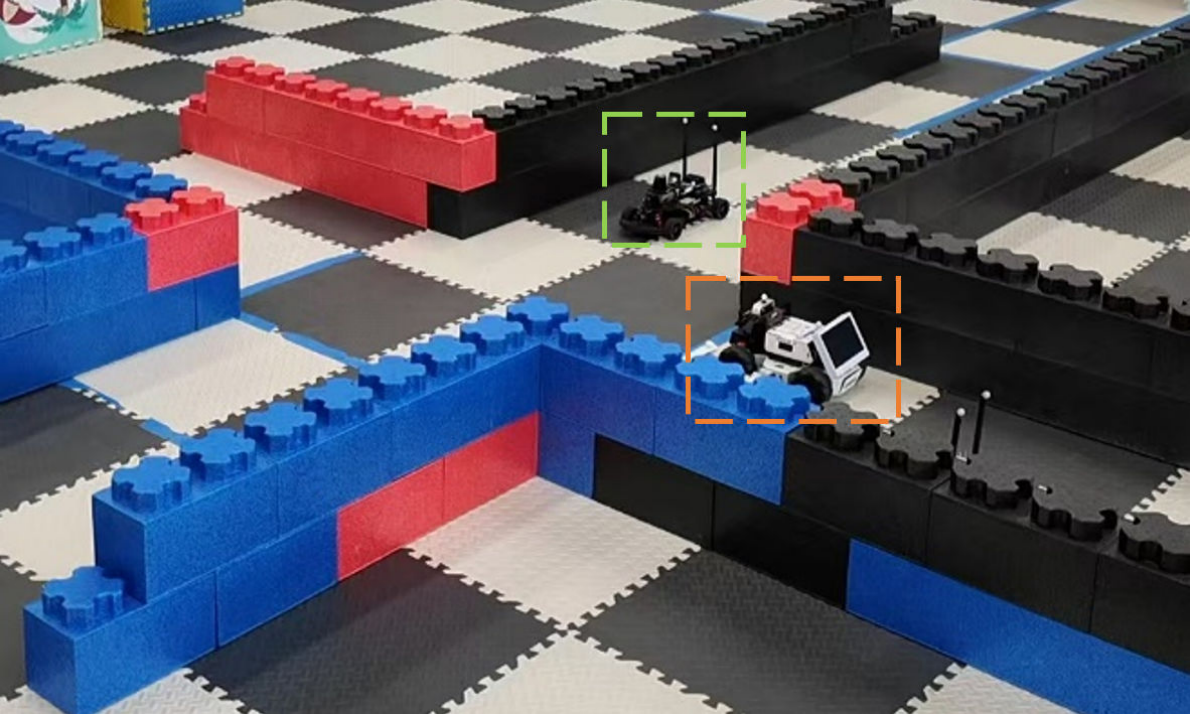} \\
\footnotesize{(b) The EV gradually reduces its speed to $0.28\,\text{m/s}$ as it approaches the intersection.} \\[5pt]
\hspace{-0mm}  \includegraphics[width=.225\textwidth]{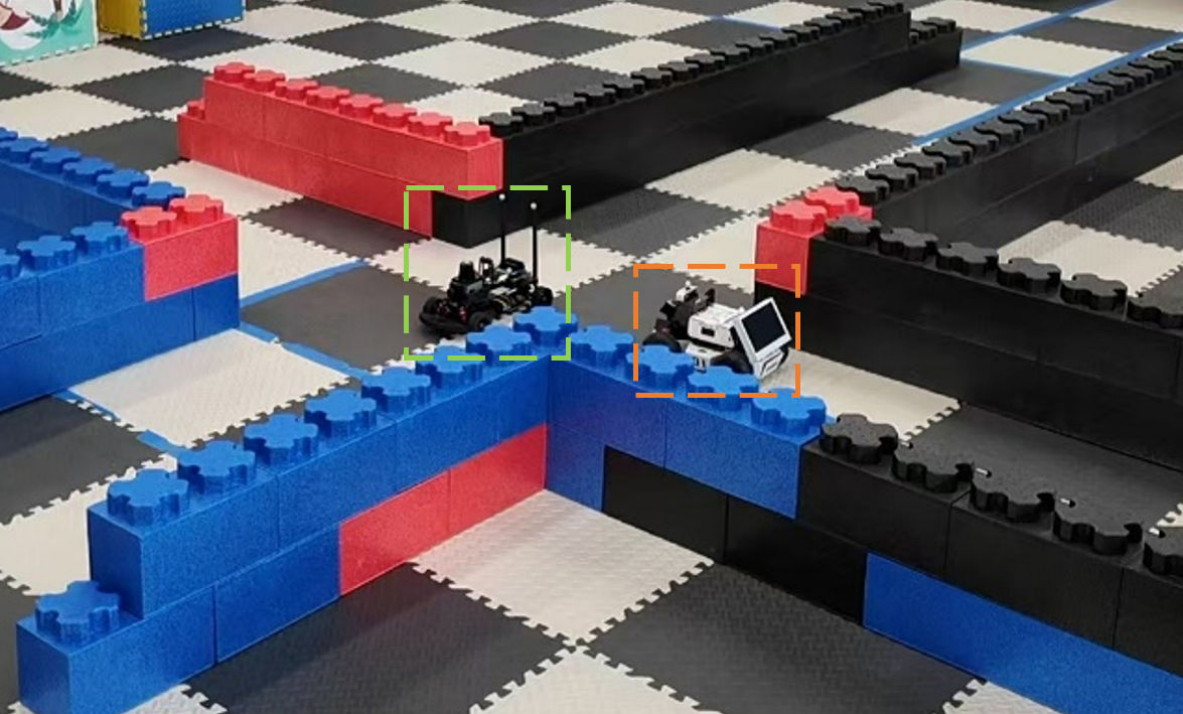}\hspace{2mm}\includegraphics[width=.225\textwidth]{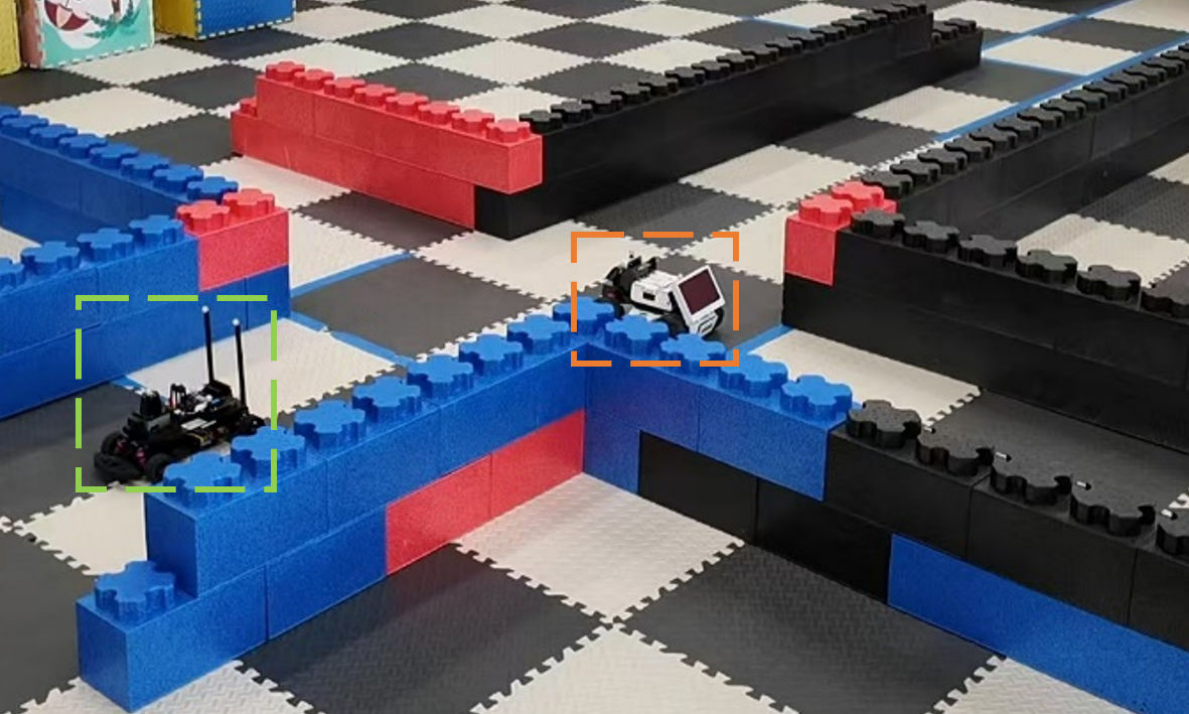}\\
\footnotesize{(c) The EV accelerates to safely pass through the intersection.}
\vspace{-2mm} 
\caption{Snapshots of a real-world experimental driving task at an occluded four-lane unsignalized intersection. The TianRacer EV (outlined with a green dashed box) navigates through the intersection, with two surrounding dynamic PVs (outlined with orange and yellow dashed boxes) maintaining their respective lanes, posing potential risks. As the EV approaches the intersection, it adjusts its speed dynamically to ensure safe navigation.} \vspace{-1mm} 
\label{fig:real_world_snapshot}
\end{figure}
\subsubsection{Experimental Setup}
To evaluate the performance of our occlusion-aware contingency planning algorithm under real-world operating conditions with limited hardware resources, we deploy the system on an autonomous TianRacer robot. The robot features an NVIDIA Jetson Xavier NX and a four-wheel configuration with Ackermann steering, simulating a scaled road vehicle, as shown in Fig.~\ref{fig:real_world_snapshot}.  Its sizes are $380\,\text{mm}$ by $210\,\text{mm}$, approximately 1:10 scale relative to a full-sized vehicle. 
{The experiment is conducted in a 1:10 scale four-way intersection, constructed with roads that are $37.5\,\text{cm}$ in width. The occlusions (visible in Fig.~\ref{fig:real_world_snapshot}) are created by static block structures forming sharp, 90-degree corners at the edge of the intersection, thereby directly obstructing the EV's visibility of the perpendicular roads.} Additionally, three PVs are deployed at the intersection, consisting of two AgileX Limo Rover robots and one TianRacer robot.

The OptiTrack\footnote{\url{https://www.optitrack.com/}} system is employed for state estimation and obstacle detection, 
providing high-precision global positioning data {(with millimeter-level accuracy post-calibration) of the robot and tracking obstacles at a frequency of 180 Hz}. Additionally, we use a moving average algorithm to estimate each robot's velocity based on the positional differences of the five most recent data points from the OptiTrack system. 

For state estimation, five reflective markers are strategically placed on the ego robot. These markers enhance the precision of the state information captured by the OptiTrack system. The computations are performed on the Ubuntu 20.04 LTS platform with ROS Noetic, with the ROS master configured to a {sampling rate of 100 Hz}, and a control frequency of 10 Hz. This setup enables the autonomous robot to perform online optimization tasks efficiently, ensuring real-time responsiveness despite the limitations imposed by the hardware resources. The maximum risk thresholds $c_{\text{th,max}}$ for the exploration and fallback trajectory are set to 4.5 and 6, respectively. The axes of the safe ellipse $l_x$ and $l_y$ are set to $60\,\text{mm}$. The maximum velocity of PVs is set to $1\,\text{m/s}$.  

    \begin{figure}[tp]
    \centering
    \subfigcapskip = -3mm
        \subfigure[ Time instant $t = 2\,\text{s}$.]{
            \label{fig:Hardware_Vel_ACC} 
            \hspace{-2mm} 
            \vspace{-1mm}
        \includegraphics[width=.45\textwidth]{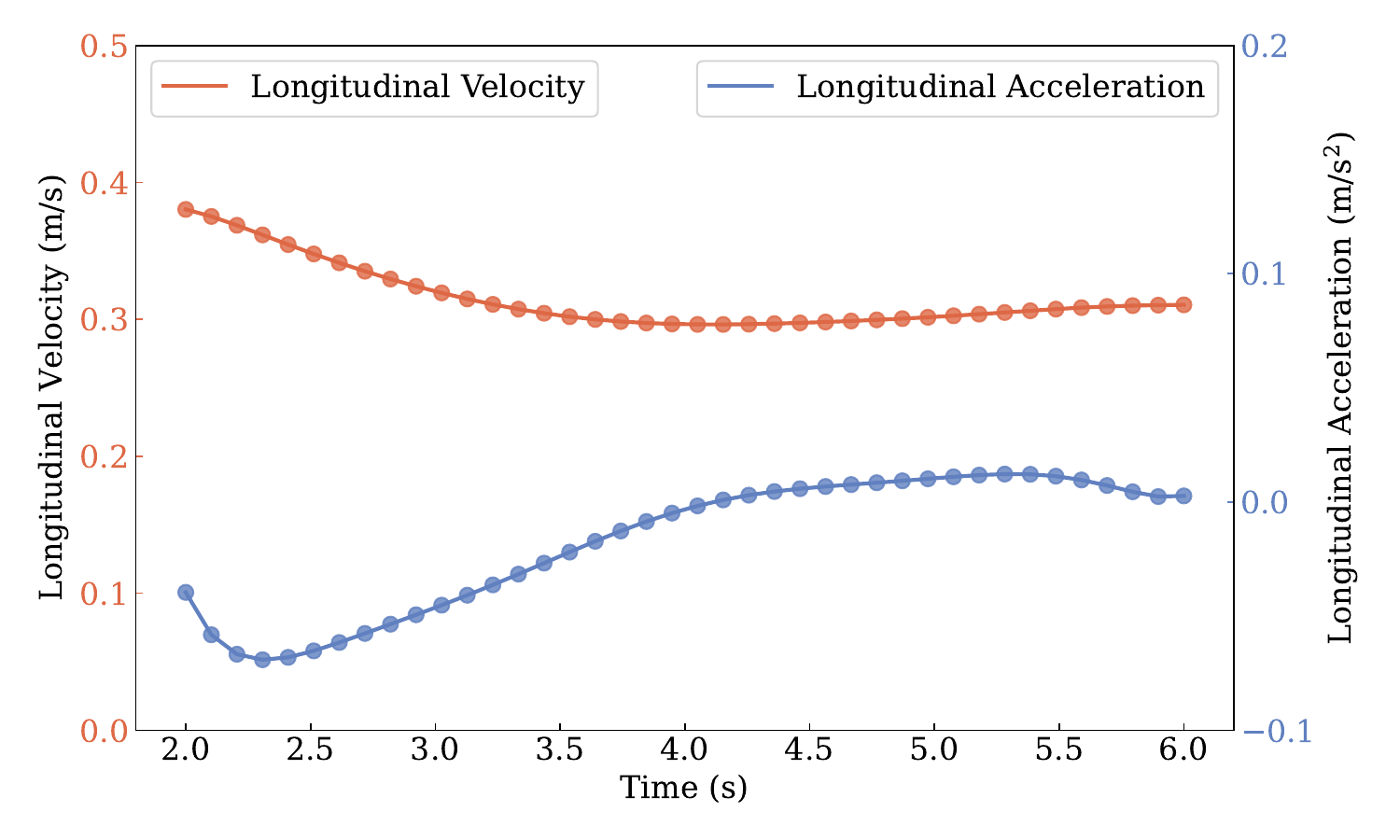}}\hspace{-2mm} 
        \subfigure[ Time instant $t = 3\,\text{s}$.]{
            \label{fig:Hardware_Vel_ACC_1}
        \includegraphics[width=.45\textwidth]{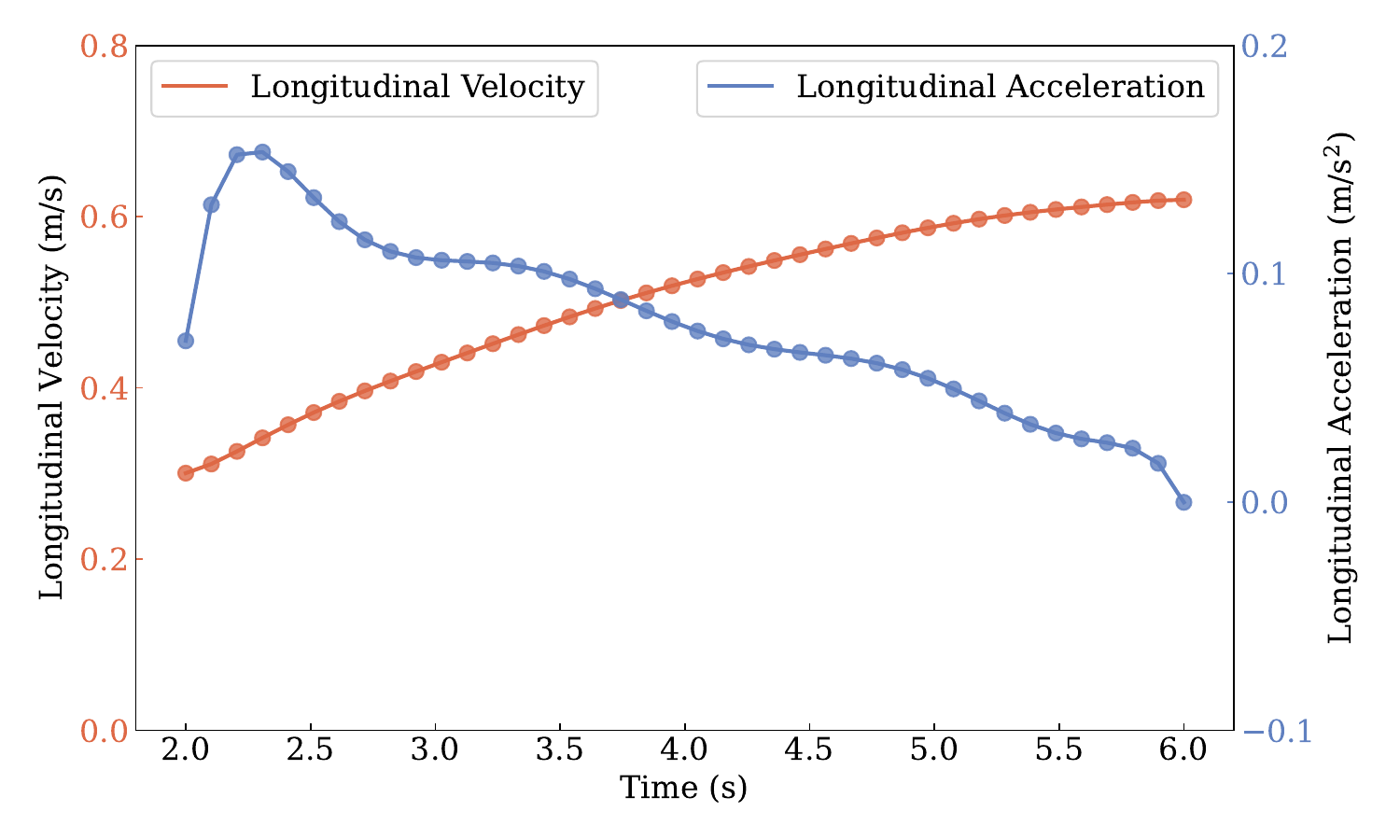}}\hspace{-2mm}
    \vspace{-1mm}
    \caption{Optimal longitudinal velocity and acceleration profiles of the selected trajectory at two different time instants. (a) The EV decelerates as it approaches the occluded intersection; (b) The EV accelerates to pass through the intersection.}	
    \label{fig:planning_vel_acc}
    \vspace{-0mm}
\end{figure}	  
 \subsubsection{Results} 
\begin{figure}[tb]
\begin{center}
\includegraphics[width=.45\textwidth]{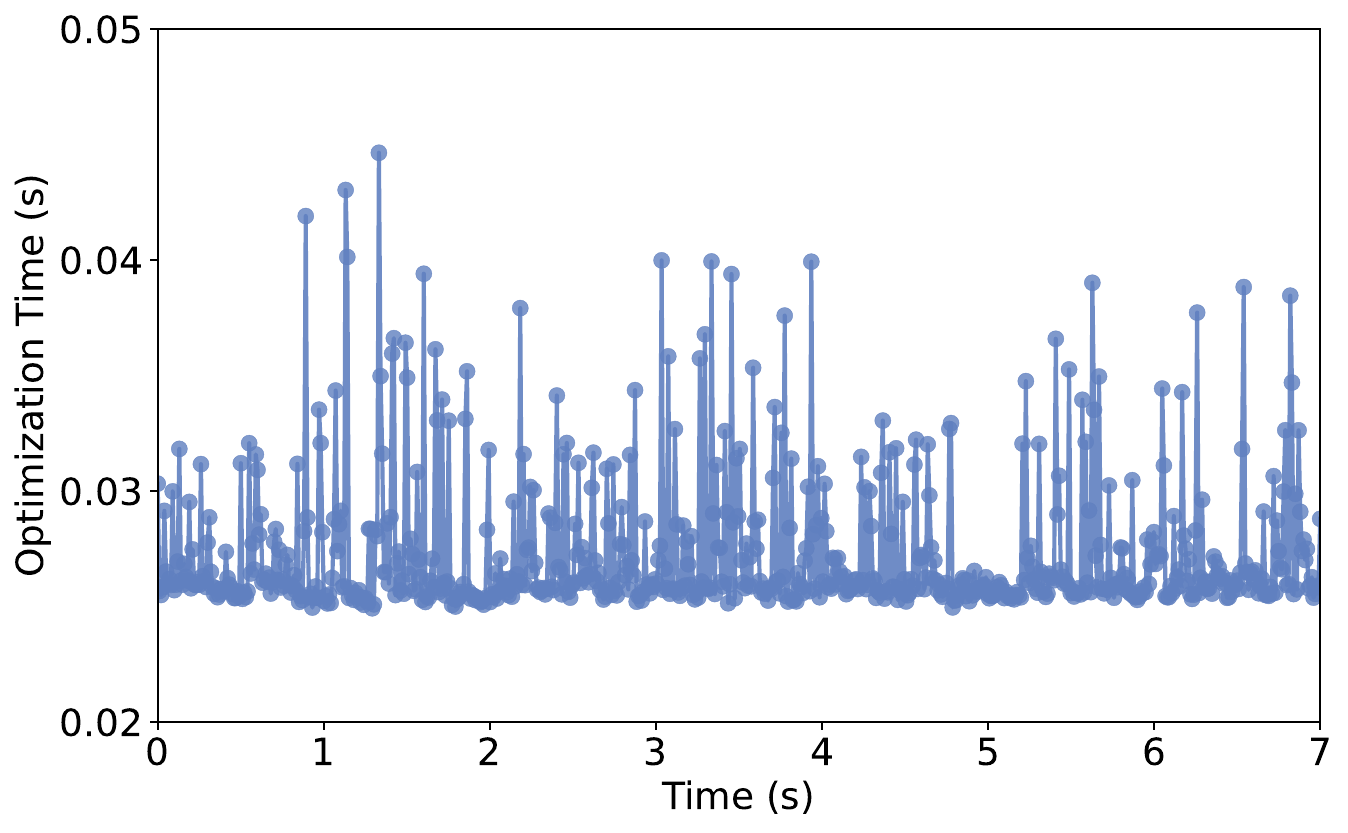}    \vspace{-2mm}
\caption{Evolution of the optimization time with a prediction step of \( N = 40 \) while interacting with three SVs. The average and maximum optimization time are $27.33\,\text{ms}$ and $44.65\,\text{ms}$, respectively.} 
\label{fig:Hardware_time}
\end{center}
\vspace{-2mm}
\end{figure}

Figure~\ref{fig:real_world_snapshot} illustrates the TianRacer navigating an occluded four-lane unsignalized intersection. In this scenario, the EV (outlined in green) encounters uncertain surrounding PVs, outlined in orange and yellow. Initially, the trajectories of these PVs do not intersect with the EV’s planned trajectory, allowing the EV to maintain a { longitudinal speed of approximately $0.5\,\text{m/s}$ along its centerline.} However, as the EV approaches the occluded intersection, it adjusts its speed slightly to avoid potential collisions with the PVs. This behavior is evident in the open-loop velocity and acceleration profiles at around $2\,\text{s}$, as shown in Fig.~\ref{fig:Hardware_Vel_ACC}.  At time $t = 2\,\text{s}$, the EV reduces its speed to $0.3\,\text{m/s}$ in response to the potential collision risk from the SVs. As the situation develops, the EV performs a sequence of replanning actions based on the dynamic environment. At $t = 3\,\text{s}$, the EV enters the intersection, with an updated perception of the surrounding environments, and accelerates to safely pass through the intersection, as shown in Figs.~\ref{fig:real_world_snapshot}(c) and~\ref{fig:Hardware_Vel_ACC_1}. The evolution of the optimization time during these maneuvers is shown in Fig.~\ref{fig:Hardware_time}. The average and maximum optimization time are $27.33\,\text{ms}$ and $ 44.65\,\text{ms}$, respectively, which demonstrates the capability of deploying our proposed algorithm in real-time applications.

These results demonstrate the effectiveness and robustness of our contingency planning algorithm in physical vehicle systems operating under occluded and obstacle-dense environments. The ability to adapt dynamically to SVs and optimize trajectory planning in real time is critical for the deployment of AVs in complex and uncertain environments.
\begin{table}[t] \centering     \scriptsize
\setlength{\abovecaptionskip}{-0mm}
\setlength{\belowcaptionskip}{-0mm}
\caption{{Statistical Results for Optimization Time (ms) vs. Number of Obstacles Over 10 Trials, Reported as Mean (Std. Dev.)}} \begin{tabular}{cccc} \hline   
Number of Obstacles  &  Average Time  & Maximum Time  & Minimum Time \\ 
\hline 
2 & 23.17 (0.18) & 40.92  (1.75)& 0.37  (0.03)\\ 
3 & 25.12  (0.25)& 39.74  (1.69)&  0.41  (0.02)\\ 
4 & 27.18  (0.26)& 43.97 (3.46)& 0.64  (0.11)\\ 
5 & 27.67  (0.37)& 46.66  (2.56)& 0.95  (0.12)\\ 
6 & 28.51  (0.38)& 44.49 (3.41) & 0.93 (0.15)\\ \hline 
\end{tabular} \label{tab:avg_time}
\end{table}
\vspace{-2mm} 

\section{Discussion}\label{sec:Discussion}

\subsection{Computational Performance} \label{subsec:computational_performance}
 To assess the computational efficiency of our proposed occlusion-aware contingency planning framework, we conduct simulations and vary the number of obstacles considered during optimization. Each simulation spanned $15\,\text{s}$, divided into 150 discrete steps. Other settings are the same as in Section~\ref{subsec:Sim_results}. The intersection scenario, depicted in Figs.~\ref{fig:sim_top-down} and \ref{fig:sim_snapshots}, consists of ten SVs traveling in two lanes and following right-hand traffic regulations within a distance range from $-70\,\text{m}$ to $70\,\text{m}$.

Table~\ref{tab:avg_time} presents the statistical results of the optimization time (in $\text{ms}$) as the considered number of obstacles increases from two to six. The data indicates that the average optimization time gradually rises with the increasing number of obstacles, moving from $23.17\,\text{ms}$ with two obstacles to $28.51\,\text{ms}$ with six obstacles. Notably, it exhibits a gradual increase with the number of obstacles, stabilizing as the obstacle count reaches five and six. Similarly, the maximum optimization time increases from $40.92\,\text{ms}$ for two obstacles to a peak of $46.66\,\text{ms}$ for five obstacles before slightly decreasing to $44.49\,\text{ms}$ for six obstacles. 
{Each ADMM iteration involves updating primal variables and dual variables to progressively approach the constraint satisfaction. As the number of obstacles increases, the size and complexity of the matrices involved in these updates also increase, extending the time required per iteration, as analyzed in Section~\ref{sec:ADMMiter}.  }  

However, when the number of obstacles exceeds four in the four-way intersection, as depicted in Figs.~\ref{fig:occ} and \ref{fig:sim_top-down}, the additional vehicles are sufficiently far from the EV, resulting in their constraints having minimal impact on the optimization process. In the ADMM framework, only active constraints, relating to closely interacting obstacles, significantly influence the optimization steps. Distant obstacles do not impose active constraints, allowing the algorithm to maintain consistent computational performance despite the increase in total obstacle count. The slight decrease in maximum optimization time at six obstacles indicates that the ADMM framework effectively isolates and processes only the relevant constraints, preventing unnecessary computational overhead from distant, non-interacting obstacles.

Overall, the computational performance analysis underscores the effectiveness and scalability of our contingency planning algorithm in managing varying numbers of obstacles within occluded driving scenarios. This efficiency is crucial for real-world applications, enabling fast computations in dense and dynamic traffic environments. \vspace{-3mm}
\subsection{{Impact of Consensus Steps} }
\label{subsec:consensus_sensitivity}
{    To evaluate the impact of the consensus horizon \(N_s\) on   computational efficiency and task performance, we conduct an ablation study across a wide range of \(N_s\) values from 3 to 20 steps under the conditions described in Section~\ref{subsec:Sim_results}.}

 {Results in Table~\ref{tab:consenus_abalation} reveal two key insights.  First, task duration and safety violations remain consistent across all $N_s$ values, demonstrating the inherent robustness of the contingency planning framework. Second, a longer consensus horizon increases the mean solve time due to the larger optimization problem. A short horizon ($N_s = 3$) achieves the lowest solve time but results in a marginally lower minimum velocity, indicating more conservative braking maneuvers. The marginally reduced velocity at \(N_s = 20\) suggests a trend toward more cautious planning with very long consensus segments.  Based on this analysis, $N_s = 5$ is selected as the optimal value. It achieves the highest minimum velocity while maintaining high computational efficiency. Future work will investigate adaptive tuning of the consensus horizon based on real-time risk confidence.  }
 
\begin{table}[t]
    \centering
    \scriptsize
    \setlength{\abovecaptionskip}{-0mm}
    \setlength{\belowcaptionskip}{-0mm}
    \caption{{Statistical Results for Ablation Study on Consensus Steps ($N_s$) Over 10 Trials, Reported as Mean (Std. Dev.)}}
    \begin{tabular}{ccccc}
        \hline
         \(N_s\) &Collision & Task Duration  & Min. Long. Vel   & Mean Solve Time  \\ 
         &  & (\text{s})  & (\text{m/s})  & (\text{ms}) \\
        \hline
        3 &\text{No}& 12.5 (0) & 1.45 (0.03) & 23.24 (0.30) \\
        \textbf{5}&\textbf{No} & \textbf{12.5 (0)} & \textbf{1.48 (0.03)} & \textbf{23.41 (0.25)} \\
        8&\text{No} & 12.5 (0) & 1.48 (0.01) & 23.44 (0.28) \\
        10&\text{No} & 12.5 (0) & 1.48 (0.01) & 23.66 (0.28) \\
        15 &\text{No}& 12.5 (0) & 1.47 (0.03) & 23.70 (0.41) \\
        20 &\text{No}& 12.5 (0) & 1.42 (0.02) & 23.78 (0.29) \\
        \hline
    \end{tabular}\vspace{-2mm}
    \label{tab:consenus_abalation}
\end{table}

\subsection{Limitations}
Although the proposed occlusion-aware contingency planner demonstrates outstanding performance in occluded intersections through both simulation and real-world experiments, it assumes that all PVs remain within their designated driving lanes. 
This limits its applicability to unstructured and highly occluded environments with dense obstacles, such as interactions with pedestrians, where obtaining reliable risk estimations using SRQ becomes challenging. Integrating advanced learning-based approaches for risk quantification~\cite{jia2024learning}  and hybrid data-driven controllers~\cite{roman2025higher}   could help overcome this limitation. 
 Additionally,  active perception methods, such as those proposed in ~\cite{higgins2021negotiating, firoozi2024occlusion}, could dynamically adjust the EV's path in unstructured scenarios to expand the perception field to enhance safety, albeit potentially compromising motion consistency.

 \vspace{-3mm}

\section{Conclusions}\label{sec:Conclusions}
In this work, we presented a novel occlusion-aware contingency planning framework. By integrating SRQ-based risk assessment into a biconvex NLP, solved efficiently via a consensus ADMM approach, our method achieved a superior balance of safety and efficiency in dynamic, occluded environments. This approach significantly improved travel efficiency, with simulations demonstrating up to a 30.17\% reduction in task duration compared to baseline methods. 
Additionally, real-world experiments on a 1:10 scale robotic platform validated the framework's practical applicability. {A detailed analysis of computational time highlighted the framework's scalability, supporting a replanning frequency exceeding 20 Hz, even as the number of obstacles increased.}
As part of future work, we will explore interactions between pedestrians and the ego agent in unstructured, occluded environments to extend the applicability of the proposed framework to more complex outdoor scenarios.
 \vspace{-2mm}
 	\bibliographystyle{IEEEtran}
	\bibliography{egbib}

\end{document}